%% file: main.tex
\pdfoutput=1
\documentclass[10pt,twocolumn,letterpaper]{article}

\usepackage[pagenumbers]{cvpr} 
\usepackage{algorithm}
\usepackage{algorithmic}
\usepackage{tikz}

\usepackage{amsmath}
\usepackage{multicol}
\usepackage{multirow}
\usepackage{amsfonts}
\usepackage{booktabs}
\usepackage{subcaption}
\usepackage{booktabs}
\usepackage{graphicx} 
\usepackage{cuted}    
\usepackage{capt-of}
\usepackage[accsupp]{axessibility}

\DeclareMathOperator*{\argmin}{arg\,min}
\input{preamble}

\usepackage{hyperref}
\definecolor{cvprblue}{rgb}{0.21,0.49,0.74}

\def\paperID{46363} 
\def\confName{CVPR}
\def\confYear{2026}

\title{DebFilter: Eradicating Biases Stashed in Value}

\author{
Seung Hyuk Lee \quad Songkuk Kim\thanks{Corresponding author}\\
School of Integrated Technology, BK21 Graduate Program in Intelligent Semiconductor Technology\\
Yonsei University\\
{\tt\small \{hyukiggle, songkuk\}@yonsei.ac.kr}
}

\begin{document}
\maketitle
\begin{strip}
    \centering
    \includegraphics[width=\linewidth]{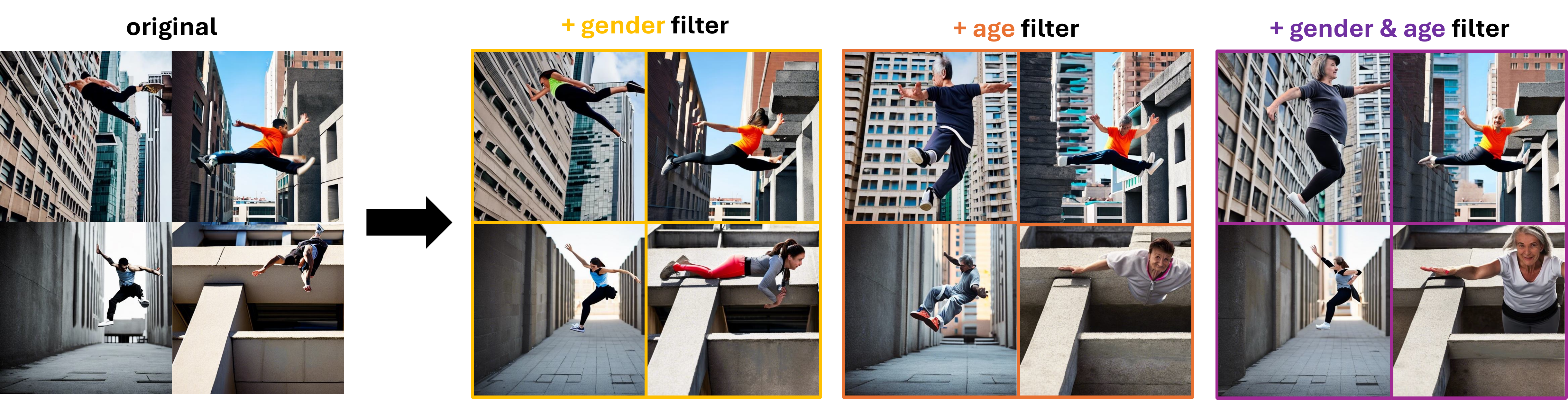}
    \captionof{figure}{Visual demonstration of ``A parkour athlete jumping between buildings'' with our flexible debiasing method \textit{DebFilter}: Selectively filtering gender (yellow), age (orange), or both concepts together (purple) in urban parkour imagery, while preserving the original action context. Manual selection of debiasing parameters enables precise control over filtered outputs. Original images shown on left.}
    \label{fig:main-parkour}
\end{strip}
\input{sec/0_abstract}    
\input{sec/1_intro}
\input{sec/2_background}
\input{sec/3_method}
\input{sec/4_experiments}
\input{sec/5_conclusion}

{
    \small
    \bibliographystyle{ieeenat_fullname}
    \bibliography{main}
}
\appendix
\input{sec/X_suppl}

\end{document}

%% file: preamble.tex
\usepackage{blindtext}

\usepackage[labelsep=period]{caption}
\usepackage[pagebackref,breaklinks,colorlinks,allcolors=cvprblue]{hyperref}
\usepackage[capitalize]{cleveref}
\crefname{section}{Sec.}{Secs.}
\Crefname{section}{Section}{Sections}
\Crefname{table}{Table}{Tables}
\crefname{table}{Tab.}{Tabs.}

%% file: sec/0_abstract.tex
\begin{abstract}

Text-to-image diffusion models, which are theoretically equivalent to score-based generative models, generate images through a multi-step denoising process guided by text embeddings extracted from pretrained vision-language models such as CLIP. 
However, these text embeddings inherently encode social and semantic biases—such as those related to gender and age—that are subsequently propagated and amplified through the guidance mechanism, along with the model’s training on large-scale datasets that are imbalanced with respect to these bias-related concepts, often leading to skewed outputs in text-to-image generation.
We propose \textbf{DebFilter}, a lightweight and training-free framework for mitigating such biases in text-to-image diffusion models. 
Observing that the model’s error prediction at each denoising step is primarily influenced by cross-attention dynamics, we introduce a bias-correction strategy that adjusts the value components within cross-attention. 
Specifically, we apply a fixed offset to the slice of guidance embedding, effectively steering the semantic direction of cross-attention values toward unbiased representations. 
This adjustment reconfigures the score landscape to produce balanced outputs while maintaining alignment with the intended text semantics.
Unlike prior approaches that rely on fine-tuning or retraining, \textbf{DebFilter} operates entirely at inference time, requiring no additional data or model updates. 
Our results demonstrate that this method effectively mitigates social biases in generated images, offering an efficient and scalable pathway toward fairer and more inclusive text-to-image generation.

\end{abstract}

%% file: sec/1_intro.tex
\section{Introduction}
\label{sec:intro}
At the core of modern text-to-image generation lies the score-based modeling framework~\cite{song2019generative,song2020score}, which learns a score function $\mathbf{s}_\theta$ that approximates the gradient of the data log-likelihood. This score function defines a vector field that progressively denoises a noisy sample, guiding it toward regions of high data density.

Contemporary diffusion-based implementations, such as Stable Diffusion~\cite{rombach2022high}, can be viewed as practical realizations of the conditional score-matching principle. While diffusion models employ a denoising U-Net~\cite{ronneberger2015u} operating through discrete noise-removal steps, score-based models describe an equivalent continuous-time process via stochastic differential equations (SDEs). Despite their different formulations, they are theoretically equivalent~\cite{song2020score}.


However, training diffusion models on large-scale image datasets, in combination with a pretrained CLIP text encoder, can lead the model to internalize and reproduce social biases present in the data~\cite{deng2009imagenet, lin2014microsoft, schuhmann2022laion, bolukbasi2016man, wang2021gender}. 
Both the visual and textual components contribute to the learned joint representation, and correlations related to gender, occupation, or other attributes may be implicitly reinforced during training. 
These biases can be further amplified by conditioning mechanisms such as classifier-free guidance~\cite{ho2022classifier}, causing even neutral prompts (e.g., ``a person who works as a doctor'') to yield skewed or stereotypical generations. 
This highlights the need for methods that mitigate such unintended biases while maintaining the semantic fidelity of the generated outputs.

Mitigating such biases is non-trivial. Existing solutions often rely on retraining or fine-tuning either the score network $\mathbf{s}_\theta$ or the text encoder, both of which demand extensive computational resources and balanced datasets—making them impractical for most real-world applications.

To address this challenge, we propose \textbf{DebFilter}, a lightweight, training-free framework for mitigating social biases in text-to-image diffusion models during inference. 
The key idea is to compute a bias-correction vector $\Delta \mathbf{c}$ and apply it to the original conditioning signal $\mathbf{c}$ throughout sampling. 
In diffusion-based generation, $\mathbf{c}$ defines the key and value representations in cross-attention layers, where values encode semantic guidance during denoising. 
Introducing $\Delta \mathbf{c}$ adjusts the semantic direction of $\mathbf{c}$, counteracting biased components in the embedding space and reshaping the score landscape toward more balanced regions of the data manifold.

Conceptually, this enables a model conditioned on a neutral prompt—e.g., \textit{``a person who works as a doctor''}—to produce more balanced outputs across attributes such as gender. 
Formally, the modified conditioning
{\small
\[
\mathbf{s}_\phi(\mathbf{z}_t \mid \mathbf{c} + \Delta \mathbf{c})
\;\approx\;
\mathbb{E}_{\mathbf{c}_{\text{explicit}} \in \mathcal{C}}[\mathbf{s}_\phi(\mathbf{z}_t \mid \mathbf{c}_{\text{explicit}})],
\]}
where $\mathcal{C}$ denotes explicit attribute-conditioned prompts (e.g., ``a man/woman who works as a doctor''). 
A single fixed offset $\Delta \mathbf{c}$ is defined per concept (e.g., gender, age, or ethnicity), serving as a universal correction direction applicable to any prompt.

As a result, even neutral prompts can yield debiased outputs, as the guidance signal is reoriented to suppress latent biases during the reverse diffusion process—without retraining or altering $\mathbf{s}_\phi$. Moreover, our method enables fine-grained, object-level control by identifying token indices corresponding to specific concepts in multi-object scenes.

Our contributions are threefold:
\textcircled{\small 1} We introduce a training-free debiasing method for Stable Diffusion that operates on error prediction by adjusting cross-attention mechanism;
\textcircled{\small 2} We demonstrate targeted, object-level control in multi-object images while preserving spatial consistency; and
\textcircled{\small 3} We present a practical and extendable approach that improves fairness and flexibility in text-to-image generation.
Together, these contributions advance the inclusivity and controllability of generative models, offering a scalable path toward socially responsible AI-driven creativity.

%% file: sec/2_background.tex
\section{Backgrounds}
\label{sec:background}

\subsection{Score-Based and Denoising Diffusion Models}
Score-based generative models and denoising diffusion models represent two closely related approaches to generative modeling based on iterative denoising. Both frameworks define a forward diffusion process that gradually adds Gaussian noise to data, and a corresponding reverse process that reconstructs clean samples from pure noise.
\paragraph{Score-Based Models.}
Score-based models~\cite{song2019generative, song2020score} learn the score function, defined as the gradient of the log-probability density of noisy data $\nabla_x \log p_t(x) $ , which indicates the direction toward higher data likelihood at each noise level \(t\). 
A neural network \(s_\phi(z_t, t)\) is trained to approximate this score under different noise scales. 
During sampling, data generation is performed by integrating a reverse-time stochastic differential equation (SDE) or its deterministic counterpart (ODE), which progressively removes noise from an initial Gaussian sample using the learned score function.
\paragraph{Denoising Diffusion Models.}
Denoising diffusion probabilistic model (DDPM)~\cite{ho2020denoising} parameterize the reverse diffusion process by directly predicting the noise component added at each timestep. The forward process is defined as:
\[
z_t = \sqrt{\bar{\alpha}_t}\,z_0 + \sqrt{1 - \bar{\alpha}_t}\,\epsilon, \quad \epsilon \sim \mathcal{N}(0, I),
\]
where \(\bar{\alpha}_t\) controls the noise level. The model $\theta$ predicts the noise term \(\hat\epsilon_{\theta}(z_{t}, c)\) and minimizes the mean squared error between the true and predicted noise. This noise-prediction formulation is mathematically equivalent to score estimation through:
\[
s_\phi(z_t, t) = -\frac{\epsilon_\theta(z_t, t)}{\sqrt{1 - \bar{\alpha}_t}}.
\]
Denoising Diffusion Implicit Model (DDIM)~\cite{song2020denoising} extends DDPM by introducing a deterministic sampling process that accelerates generation while preserving the same marginal data distribution. Unlike the stochastic nature of DDPM, DDIM removes noise injection at each denoising step and formulates a non-Markovian deterministic mapping between noise and data, enabling faster and more controllable sampling.

\subsection{Stable Diffusion and Guidance via CLIP}

Stable Diffusion is a powerful latent diffusion model (LDM) designed to generate high-quality images efficiently by operating in a compressed latent space rather than pixel space. This is made possible by an autoencoder architecture, where a variational autoencoder (VAE)~\cite{kingma2013auto} compresses the input image $\mathbf{x} \in \mathbb{R}^{H \times W \times 3}$ into a latent representation $\mathbf{z} \in \mathbb{R}^{h \times w \times c}$ such that $h, w = H/f, W/f$ with compression factor $f$, i.e., $\mathbf{z} = \mathcal{E}(\mathbf{x})$. The diffusion process then operates on this latent space:
\[
q(\mathbf{z}_t|\mathbf{z}_0) = \mathcal{N}(\mathbf{z}_t; \sqrt{\bar{\alpha}_t}\mathbf{z}_0, (1-\bar{\alpha}_t)\mathbf{I})
\]
where $t \in [0,1]$ is the diffusion timestep and $\bar{\alpha}_t$ follows a cosine noise schedule. 
\begin{table}[htb]
    \centering
    \footnotesize
    \resizebox{\columnwidth}{!}{%
        \begin{tabular}{|c|c|c|c|c|c|c|c|c|c|}
            \hline
            1 & 2 & 3 & 4 & 5 & 6 & 7 & 8 & ... & 77 \\
            \hline
            \rule{0pt}{2.5ex}\texttt{<sos>} & A & \textbf{CEO} & in & an & office & \texttt{<eos>} & \texttt{<eos>} & \texttt{...} & \texttt{<eos>} \\
            \hline
        \end{tabular}%
    }
    \caption{Example of tokenized sequence of the prompt ``A CEO in an office'' with corresponding token indices. Padding tokens are \texttt{<eos>}.}
    \label{tab:tokens}
\end{table}
A core innovation in Stable Diffusion is its use of Contrastive Language–Image Pretraining (CLIP)~\cite{radford2021learning,cherti2023openclip} for conditioning image generation on textual prompts. CLIP is a multimodal model that learns visual concepts through natural language supervision by aligning image and text embeddings via contrastive learning. It processes input text into a sequence of 77 fixed-length token embeddings that encapsulate semantic content as can be seen in Table~\ref{tab:tokens}. These embeddings, denoted by $\mathbf{c}$, are used as conditioning guidance in the denoising U-Net $\theta$, enabling text-guided image synthesis with predicting \(\epsilon_\theta(\mathbf{z}_t, t, \mathbf{c})\). During inference, the denoised latent $\mathbf{z}$ is decoded to the final image using the decoder $\mathcal{D}$ as \(\mathbf{x} = \mathcal{D}(\mathbf{z})\).

Stable Diffusion employs classifier-free guidance~\cite{ho2022classifier} to steer generations more strongly toward the prompt semantics. Instead of relying on an external classifier~\cite{dhariwal2021diffusion}, the model is trained to predict noise both conditionally and unconditionally. Specifically, the noise prediction is computed as: 
\begin{equation} 
\hat{\epsilon}_\theta(z_t, c) = (1 + w)\epsilon_\theta(z_t, c) - w \epsilon_\theta(z_t, \phi)
\end{equation}
where $w$ is a guidance weight that controls the strength of the conditioning. This approach avoids external classifiers and improves semantic alignment while preserving diversity.

Crucially, textual semantics are injected via cross-attention layers within the denoising U-Net, where visual features (queries) attend over the textual features (keys and values). These mechanisms are central to how semantic information from the prompt influences the image synthesis. Our method, \textbf{DebFilter}, focuses specifically on the value component of these layers to mitigate bias.

\subsection{Mitigating Bias in Generative Models}
Efforts to mitigate bias in generative models often face a trade-off between controllability and unintended semantic distortion. 
ENTIGEN~\cite{bansal2022well} introduces ethical constraints through prompt manipulation, but this can alter the original meaning of the text and shift image semantics. 
ITI-GEN~\cite{zhang2023iti} addresses fairness by learning category-specific tokens, yet its reliance on numerous reference images per attribute limits practicality in real-world applications. 

Other approaches directly intervene in the model’s internal mechanisms. 
TIME~\cite{orgad2023editing} modifies cross-attention layers to edit concepts but may cause ripple effects that unintentionally alter unrelated features. 
UCE~\cite{gandikota2024unified} improves on this by enabling multi-concept editing while preserving selected attributes, though it still requires manually defined attribute lists. 
MIST~\cite{yesiltepe2024mist} disentangles attributes via the \texttt{<eos>} token, allowing more localized cross-attention updates and finer bias control. While MIST targeted on role of $<eos>$ tokens, we differ by targeting prompt-word embeddings directly instead of padding.
Nevertheless, these methods often remain limited to single-concept modification or demand additional training for each adjustment. 

For vision–language models, debiasing has centered on refining learned representations. 
DeAR~\cite{seth2023dear} applies adversarial training to modify image features but is sensitive to hyperparameter selection. 
CLIP-clip~\cite{wang2021gender} reduces bias by pruning embedding dimensions, at the risk of semantic degradation. 
SFID~\cite{jung2024unified} combines feature pruning with confidence-based imputation to balance fidelity and fairness, yet still depends on curated debiasing datasets and accurate identification of biased components.

%% file: sec/3_method.tex
\section{Proposed Method}
\label{method}
The direction of image generation is governed by the model’s denoising prediction $\hat{\epsilon}_{\theta}(z_{t}, c)$. Because the conditioning embedding $c$ defines the structure of the score-function landscape, it implicitly determines the generative trajectory. Since $\hat{\epsilon}_{\theta}(z_{t}, c)$ is biased, we modify the direction of the score landscape by adding a correction term $\Delta\epsilon$. The appropriate correction direction is unknown, so we introduce a target score $\hat{\epsilon}_{\theta}(z_{t}, c')$ and enforce the relation
$\hat{\epsilon}_{\theta}(z_{t}, c) + \Delta\epsilon \approx \hat{\epsilon}_{\theta}(z_{t}, c')$, 
where $c'$ is the embedding of an explicit prompt chosen to invert the relevant bias attribute. 
A key observation is that $\hat{\epsilon}_{\theta}(z_{t}, c)$ can be expressed as a linear function of the cross-attention(CA) output, which enables computation of $\Delta\epsilon$ directly from the difference $\Delta\mathrm{CA}$.

\begin{figure*}[!htbp]
    \centering
    \begin{subfigure}{0.55\linewidth}
        \includegraphics[width=\linewidth]{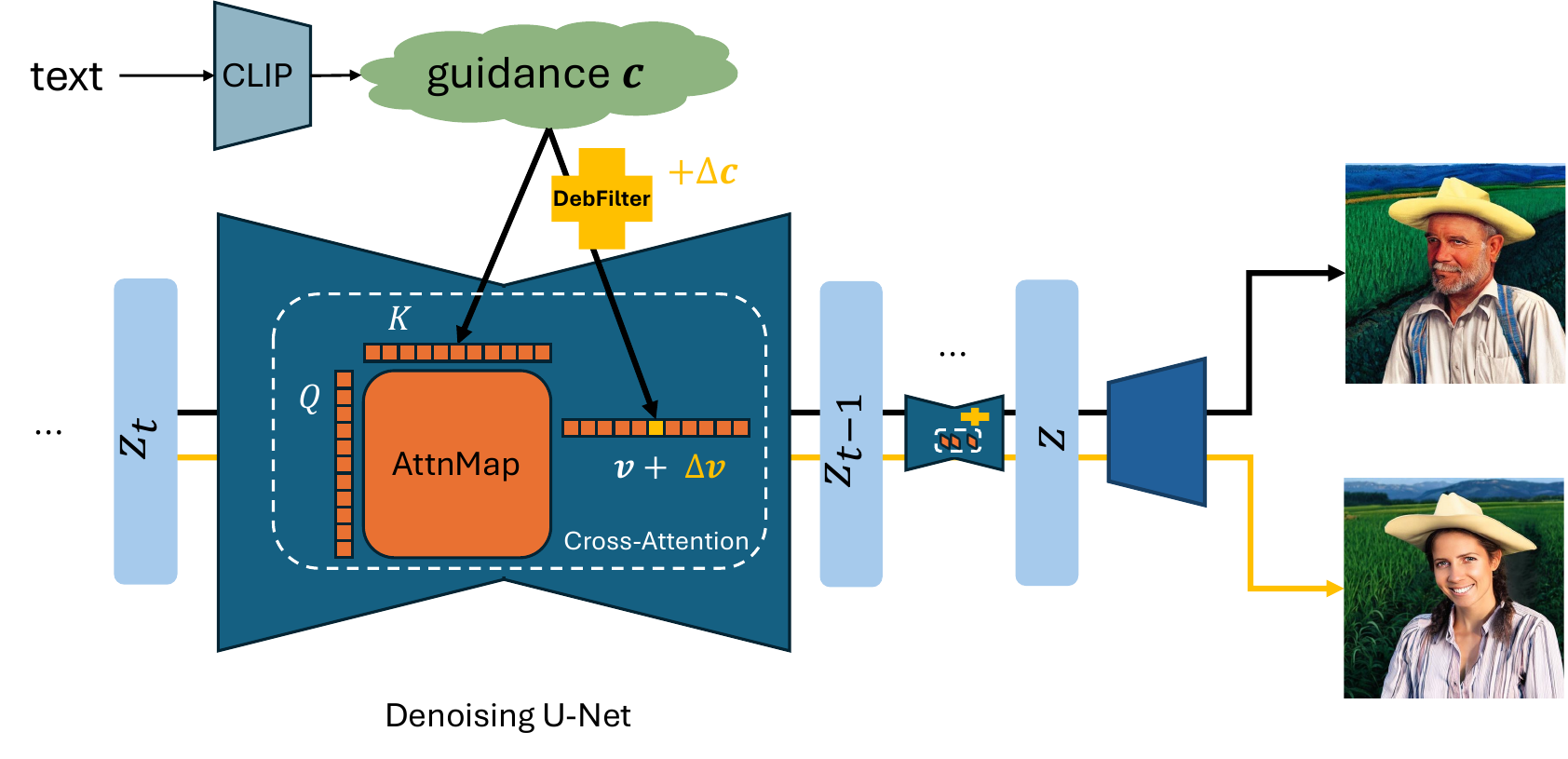}
        \caption{}
        \label{fig:concept}
    \end{subfigure}
    \hspace{10pt}
    \begin{subfigure}{0.4\linewidth}
        \includegraphics[width=\linewidth]{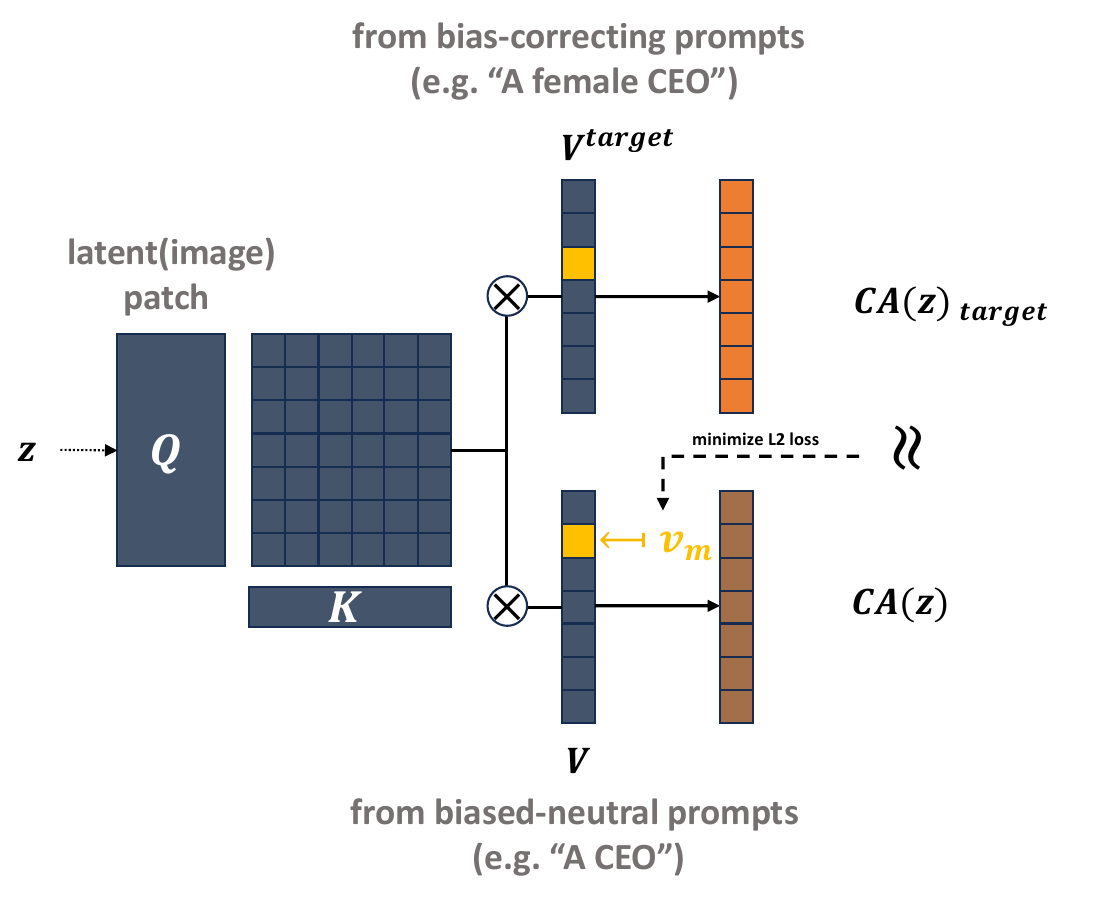}
        \caption{}
        \label{fig:concept-little}
    \end{subfigure}
    \caption{(a) Illustration of the \textit{DebFilter} architecture for debiasing diffusion models. The diagram shows a denoising U-Net pipeline with cross-attention mechanism where text prompts are processed through CLIP text encoding. \textit{DebFilter} is integrated into this pipeline to mitigate biases during the image generation process by specifically filtering the value component of the cross-attention mechanism. (b) Cross-Attention with modified $\textbf{v}_{m}^{gender}$ using linear regression. Only the value corresponding to the index of the target subject for debiasing is optimized.}
    \label{fig:cross-attention}
\end{figure*}

\subsection{Cross-Attention Alignment For Matching Score Function}
The output $\hat{\epsilon}_{\theta}(z_{t}, c)$ is produced by the denoising U\text{-}Net, which contains stacked self\text{-}attention (SA) and cross\text{-}attention (CA) layers~\cite{vaswani2017attention}. Since the conditioning embedding $c$ enters the model only through CA, we can intervene directly on this component. Owing to the alternating SA--CA structure, modifying CA also affects subsequent SA layers. By aligning the CA outputs under $c$ and its counterpart $c'$, we adjust the denoising prediction without altering any model weights.

Formally, the denoising prediction can be expressed as a function of the intermediate cross-attention activations: 
$\hat{\epsilon}_{\theta}(z_t, c) = f_{\theta}\big(\mathrm{CA}(z_t, c)\big)$,
where $\mathrm{CA}(z_t, c)$ denotes the aggregated cross-attention output across all layers. 
Conventional approaches may update model parameters from $\theta$ to $\theta'$ to achieve $f_{\theta'}(\mathrm{CA}(z_t, c)) \approx f_{\theta}(\mathrm{CA}(z_t, c'))$, 
requiring fine-tuning with explicit attribute prompts. 
In contrast, we directly modify the conditioning input by constructing $\tilde{c} = c + \Delta c$ such that
{\small
\begin{equation}
\mathrm{CA}(z_t,\tilde{c}) = \mathrm{CA}(z_t,c)+\Delta\mathrm{CA} \approx \mathrm{CA}(z_t,c'),
\end{equation}
}
which yields the corresponding denoising behavior
{\small
\begin{equation}
\hat{\epsilon}_{\theta}(z_t,\tilde{c}) \approx \hat{\epsilon}_{\theta}(z_t,c)+\Delta\epsilon \approx \hat{\epsilon}_{\theta}(z_t,c').
\end{equation}
}
This adjustment reproduces the effect of retraining while keeping all model parameters fixed, enabling efficient and training-free bias correction during inference. In this way, aligning cross-attention activations through $\tilde{\mathbf{c}}$ effectively aligns the model’s denoising trajectory without modifying $\theta$ itself.

In the cross-attention operation, queries are derived from the latent variable (noise) $Z$, while keys and values are obtained from the conditioning embedding $\mathbf{c}$ through linear projections. 
This process can be formally expressed as:
{\small
\begin{equation}
\begin{aligned}
    \mathrm{CA}^{h}(i) &= \sum_{j=1}^{77} A^{h}_{i,j} \cdot (W^{h}_v \mathbf{c}_{j}) = \sum_{j=1}^{77} A^{h}_{i,j} \cdot \mathbf{v}^{h}_j, \\
    \text{where} \quad 
    A^{h} &= \mathrm{Softmax}\left( \frac{W^{h}_q Z \cdot (W^{h}_k \mathbf{c})^\top}{\sqrt{d_k}} \right),
\end{aligned}
\label{eq:summation}
\end{equation}
}
where $\mathrm{CA}^{h}(i)$ denotes the cross-attention output for the $i$-th latent token in head $h$, and $A^{h}_{i,j}$ represents the attention weight between latent token $i$ and conditioning token $j$.

Since attention weights can be interpreted as coefficients that weight the value vectors $\mathbf{v}^{h}_{j}$, modifying a particular value corresponding to a biased token provides a direct way to correct the attention outcome. 
Rewriting Eq.~\ref{eq:summation} to isolate a specific token index $m$ yields:
{\small
\begin{equation}
\mathrm{CA}^{h}_{\text{source}}(i) 
= \sum_{j=1}^{m-1} A^{h}_{i,j} \cdot \mathbf{v}^{h}_{j} 
+ A^{h}_{i,m} \cdot \mathbf{v}^{h}_{m} 
+ \sum_{j=m+1}^{77} A^{h}_{i,j} \cdot \mathbf{v}^{h}_{j}.
\label{eq:source}
\end{equation}
}

To compute $\Delta\mathrm{CA}$, we subtract the target cross-attention output $\mathrm{CA}^{h}_{\text{target}}(i)$, derived from $c'$, from the source output $\mathrm{CA}^{h}_{\text{source}}(i)$. We then adjust the corresponding value vector $\mathbf{v}^{h}_{m}$ by solving the following least-squares optimization:
{\small
\begin{equation}
\textbf{v}_{m}^{h}
= \argmin_v 
\sum_{i} 
\left(
A_{i, m}^{h} \cdot v
-
\left(
\text{CA}_{\text{target}}^{h}
-
\sum_{j \ne m} A_{i, j}^{h} \cdot \textbf{v}_{j}^{h}
\right)
\right)^2
\label{eq:regression}
\end{equation}
}
This formulation yields the optimal correction to the biased value vector, aligning the attention output of $c$ with that of $c'$ and mitigating prediction bias without retraining.

Since $\mathbf{v}^{h}_{m}$ is linearly projected from the conditioning embedding $\mathbf{c}$, 
the optimization in Eq.~\ref{eq:regression} can be solved using standard linear regression techniques. $\textbf{v}_m$ of Figure~\ref{fig:concept-little} illustrates how newly computed $\mathbf{v}^{h}_{m}$ matches $\mathrm{CA}^{h}_{\text{source}}(i)$ to $\mathrm{CA}^{h}_{\text{target}}(i)$.

\subsection{Unified Guidance Estimation Using Aligned Multi-Layer Cross-Attention}
\label{method:new_g}
In the denoising U-Net, each denoising step comprises multiple attention blocks with varying depths, 
and each block differs in the number of attention heads and feature dimensions(specifically described in Supplementary Materials). 
Therefore, for every depth, we independently estimate the target value vector $\hat{\mathbf{v}}^{h}_{m}$ 
that aligns the cross-attention output with the target. 


Since $\hat{\mathbf{v}}^{h}_{m}$ is a linear projection of $\mathbf{c}_m$, 
we aggregate all per-depth estimates $\hat{\mathbf{v}}_{m}^{\,l,h}$ and their projection weights $W_v^{l,h}$, and obtain the unified debiased embedding $\mathbf{c}^{\text{new}}_m$ via a weighted least-squares regression:
\begin{equation}
\mathbf{c}^{\text{new}}_m
= \argmin_{c}
\sum_{l,h,i}
\bigl\|
W_v^{l,h} \mathbf{c} - \hat{\mathbf{v}}_{m}^{\,l,h}
\bigr\|_2^2
\end{equation}
The updated embedding $\mathbf{c}^{\text{new}}_m$ produces reprojected values $W_v^{l,h} \mathbf{c}^{\text{new}}_m$ that closely match the targets $\hat{\mathbf{v}}_{m}^{\,l,h}$ across layers and heads. Configuration details of U-Net structure are illustrated in supplementary materials.

Because the early denoising steps primarily determine the coarse structure and high-level attributes of the image, we compute and average $\mathbf{c}^{\text{new}}_m$ only over the first two denoising steps. This early-stage estimation captures the dominant semantic bias without being influenced by later refinement processes focused on texture or style.

\subsection{DebFilter: Deriving a Stable Debiasing Guidance from Cross-Prompt Value Offsets}
Using the updated embedding $\mathbf{c}^{\text{new}}_m$ reduces the model’s error prediction with the target, allowing the generated output to approximate a debiased result. However, computing $\mathbf{c}^{\text{new}}_m$ for every prompt is inefficient and not scalable, as it requires a reference target at each inference step. To address this, we store the offset 
$\Delta \mathbf{c}_m = \mathbf{c}^{\text{new}}_m - \mathbf{c}_m$ and average it across multiple prompts:
\begin{equation}
\Delta \overline{\mathbf{c}} = \mathbb{E}_{p \in \mathcal{P}}[\Delta \mathbf{c}^{p}_m]
\end{equation}

An offset derived from a single prompt tends to generalize poorly, as the degree of adjustment is highly sensitive to the chosen reference text. To assess cross-prompt consistency, we aggregate the value representations across all heads and cross-attention layers for tokens such as ``female'' and ``doctor'' in the composite prompt ``female doctor.'' 
The resulting composite vector is then compared to that of ``doctor'' alone, revealing a high cosine similarity between $\Delta \mathbf{v}_{\text{doctor}}^{\text{debias}}$ and $\Delta \mathbf{v}_{\text{firefighter}}^{\text{debias}}$. This finding indicates that the transformation is semantically consistent across occupational concepts, underscoring the importance of averaging offsets from multiple prompts to achieve a generalizable debiasing direction. Similarity comparison is illustrated in supplementary materials.

We define the averaged offset $\Delta \overline{\mathbf{c}}$ as \textbf{DebFilter}, a stable debiasing direction in the embedding space. Applying \textbf{DebFilter} reorients the conditioning embedding $\mathbf{c}$, reducing latent bias while preserving the prompt’s intended meaning. For instance, when used with a neutral prompt such as ``a person who works as a doctor,'' \textbf{DebFilter} yields more balanced gender representations, with similar improvements for prompts like ``a person who works as a firefighter.''

To construct \textbf{DebFilter}, we proceed as follows. In the first two denoising steps, an estimate $\hat{\mathbf{v}}_m$ is collected from each of the 16 cross-attention layers. For each $\hat{\mathbf{v}}_m$, we identify its corresponding index in the guidance representation and compute its difference from the original guidance. Averaging these differences across multiple prompts produces the final offset.

Once computed, the same \textbf{DebFilter} is applied to the conditioning embedding $\mathbf{c}$ across all denoising steps as illustrated in Figure~\ref{fig:concept}. With only a small perturbation to the guidance embedding, \textbf{DebFilter} reshapes the diffusion trajectory, steering the score landscape toward more balanced semantic regions. This leads neutral prompts to produce more inclusive outputs while preserving semantic fidelity.

In contrast to prior approaches~\cite{gandikota2024unified, orgad2023editing, zhang2023iti}, 
our method requires neither additional training nor external data, and introduces no architectural modifications for particular cases. 

%% file: sec/4_experiments.tex
\section{Experiments}
\label{experiments}
\paragraph{Setups}
We conduct our experiments using Stable Diffusion v2.1 (SD) as the baseline generative model. \textbf{DebFilter} is applied uniformly across all denoising steps to ensure consistent intervention.

We evaluated \textbf{DebFilter} by averaging offsets from two to six occupation-related prompts for gender bias mitigation. Averaging three occupations produced the lowest LPIPS score, offering the most stable debiasing effect with minimal distortion. Thus, offsets from \textit{CEO}, \textit{firefighter}, and \textit{driver} were used to construct \textbf{DebFilter}, and these professions were excluded from later evaluations. Additional averaging results are provided in the supplementary materials.

For age bias mitigation, \textbf{DebFilter} was constructed using a single prompt because age-related attributes (e.g., skin texture) produced nearly identical offsets across prompts. A single prompt therefore sufficed to obtain a stable and effective age-related \textbf{DebFilter}.

\subsection{Gender Bias Mitigation}
\label{sec:exp-gender}
With the aid of WinoBias~\cite{zhao2018gender} and DALL-Eval~\cite{Cho2023DallEval}, we select a total of 87 occupations, consisting of 58 male-dominant occupations—predominantly represented by men—and 29 female-dominant occupations—mostly represented by women—for image generation. While WinoBias categorizes certain occupations as socially male- or female-dominated, we observe that the gender distribution in generated outputs is often inverted. To investigate this discrepancy, we generate images for each occupation using a neutral prompt such as ``A person who works as $[occupation]$'' with vanilla SD model. We then estimate gender distributions using CLIP (ViT-B/32)~\cite{radford2021learning}-based few-shot classification. Based on these results, we conduct a series of debiasing experiments under various conditions.

\begin{figure}[tbp]
    \centering
    \includegraphics[width=0.88\linewidth]{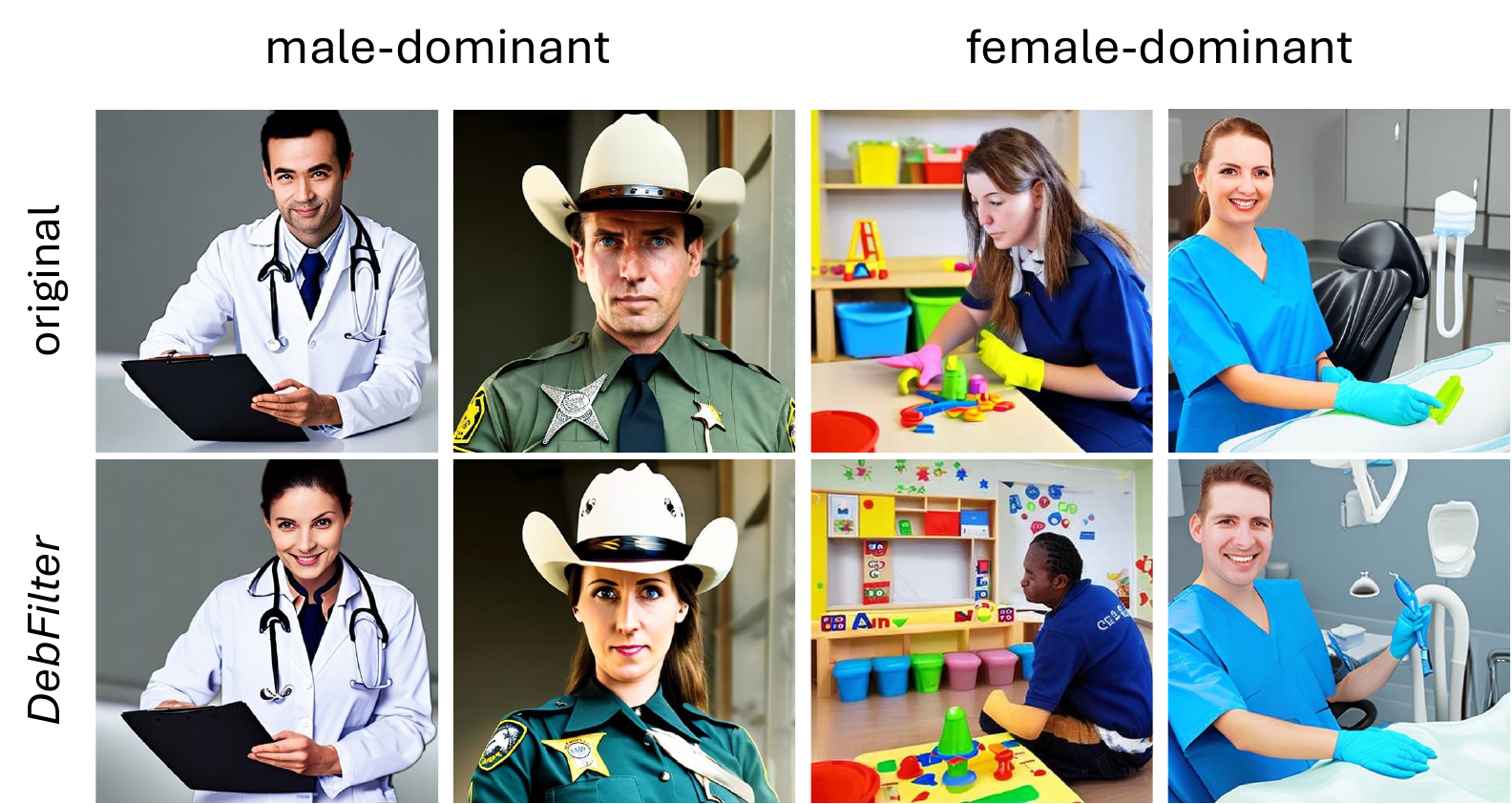}
    \caption{Examples of \textit{doctor}, \textit{sheriff}, \textit{teacher} and \textit{dentist} using \textit{DebFilter} for gender debiasing.}
    \label{fig:m2f}
\end{figure}

We demonstrate the effectiveness of our method in mitigating bias through four quantitative evaluation metrics. Figure~\ref{fig:m2f} illustrates that \textbf{DebFilter} effectively modifies gender-related attributes while maintaining the image’s overall structure. More qualitative examples are provided in the supplementary materials.
\subsubsection{Bias Ratio}
Following prior work such as TIME~\cite{orgad2023editing} and UCE~\cite{gandikota2024unified}, we calculate the proportion of generated female-presenting figures for each profession, denoted as  $F_p \in [0, 100]$, to assess gender inequality in the model's representation of professions. For each prompt, we generate 120 images and use CLIP to classify gender in the generated outputs. 

We then quantify the gender bias for each profession as the normalized absolute deviation of $F_p$ from the ideal 50\% balance, defined as
{\small
\begin{equation}
    \Delta_p = \frac{|F_p - 50|}{50}.
\end{equation}
}
To obtain an overall gender bias metric for the model, we average \( \Delta_p \) across all professions: \(\Delta = \frac{1}{|P|} \sum_{p \in P} \Delta_p\) where \( P \) is the set of all evaluated professions. A lower \( \Delta = 0 \) indicates a more balanced results and a perfectly unbiased model would yield \( \Delta = 0 \).

With analyzing the gender distribution from the vanilla generation and then adjusting the debiasing ratio of \textbf{DebFilter} accordingly, it is possible to generate outputs with gender proportions that result in a \( \Delta \) value closer to zero across occupations.

\begin{figure}[htp]
    \centering
    \includegraphics[width=0.9\linewidth]{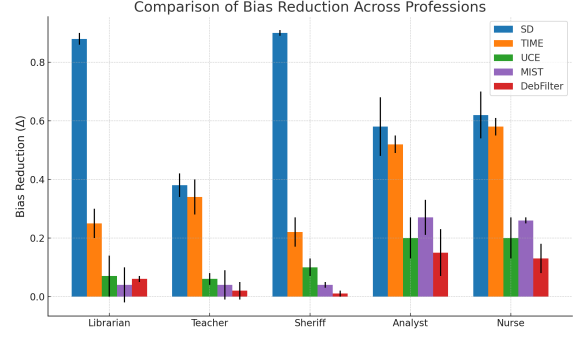}
    \caption{Comparison of bias reduction(\(\Delta\)) across professions for different debiasing models. Lower \(\Delta\) implies more balanced results.}
    \label{tab:delta-comparison}
\end{figure}

As shown in Figure~\ref{tab:delta-comparison}, \textbf{DebFilter} outperforms TIME, UCE, and MIST across selected professions evaluated under the UCE protocol. 

\begin{table*}[ht]
\centering
\vspace{0.5em}
\small
\begin{minipage}[t]{0.31\textwidth}
    \centering
    (a) Male-dominant occupations
    \vspace{0.5em}
    \begin{tabular}{@{}lccc@{}}
        \toprule
        \textbf{Occupation} & \(\Delta_{org}\downarrow\) & \(\Delta_{deb}\downarrow\) & TS \\
        \midrule
        scientist    & 0.783 & 0.067 & 0.40 \\
        engineer     & 0.883 & 0.050 & 0.49 \\
        doctor       & 0.617 & 0.250 & 0.53 \\
        sheriff      & 0.492 & 0.017 & 0.46 \\
        journalist   & 0.783 & 0.017 & 0.43 \\
        electrician  & 0.983 & 0.033 & 0.48 \\
        plumber      & 0.983 & 0.033 & 0.48 \\
        mechanic     & 1.000 & 0.050 & 0.48 \\
        cook         & 0.817 & 0.117 & 0.51 \\
        manager      & 0.683 & 0.017 & 0.42 \\
        \bottomrule
    \end{tabular}
\end{minipage}
\hspace{0.02\textwidth}
\begin{minipage}[t]{0.31\textwidth}
    \centering
    (b) Female-dominant occupations
    \vspace{0.5em}
    \begin{tabular}{@{}lccc@{}}
        \toprule
        \textbf{Occupation} & \(\Delta_{org}\downarrow\) & \(\Delta_{deb}\downarrow\) & TS \\
        \midrule
        nurse             & 0.950 & 0.050 & 0.45 \\
        dentist           & 0.317 & 0.017 & 0.49 \\
        caretaker         & 0.467 & 0.200 & 0.33 \\
        hairdresser       & 0.317 & 0.417 & 0.38 \\
        translator        & 0.883 & 0.067 & 0.28 \\
        civil servant     & 0.183 & 0.117 & 0.42 \\
        teacher           & 0.383 & 0.233 & 0.37 \\
        police officer    & 0.417 & 0.133 & 0.38 \\
        personal assistant& 0.817 & 0.017 & 0.41 \\
        makeup artist     & 1.000 & 0.383 & 0.30 \\
        \bottomrule
    \end{tabular}
\end{minipage}
\hspace{0.02\textwidth}
\begin{minipage}[t]{0.31\textwidth}
    \centering
    (c) \textsc{CLIPScore} comparison
    \vspace{0.5em}
    \begin{tabular}{@{}lcc@{}}
        \toprule
        \textbf{Occupation} & \textbf{Vanilla} & \textbf{DebFilter} \\
        \midrule
        Teacher     & 26.30 & \textbf{26.51} \\
        Librarian   & \textbf{29.10} & 28.85 \\
        Lawyer      & 27.53 & \textbf{28.17} \\
        Doctor      & \textbf{29.09} & 28.90 \\
        Sheriff     & \textbf{30.63} & 30.43 \\
        Executive   & 26.68 & \textbf{27.24} \\
        Receptionist& \textbf{28.75} & 28.63 \\
        Housekeeper & 31.19 & \textbf{31.34} \\
        \midrule
        Mean        & 28.65 & \textbf{28.76} \\
        \bottomrule
    \end{tabular}
\end{minipage}
\caption{Debiasing performance across occupations. (a) Male-dominant and (b) female-dominant occupations show substantial reductions in occupational bias after applying \textbf{DebFilter}. (c) \textsc{CLIPScore
} comparison indicates that \textbf{DebFilter} achieves comparable or improved alignment quality relative to the vanilla model.}
\label{tab:occupation_all}
\end{table*}


To examine the applicability of our approach to various scenarios, we conducted an experiment by applying the $\Delta \mathbf{c}$ obtained from the male-dominant examples with the opposite sign. Table~\ref{tab:occupation_all}(b) illustrates that the $\Delta \mathbf{c}$ obtained from male-to-female debiasing can produce a comparable effect when applied in the reverse direction (i.e., as a female-to-male filter by simply inverting the direction). This demonstrates that \textbf{DebFilter} can compute in the opposite direction simply by changing the sign similar to vector operations. Consequently, there is no need to separately compute and store $\Delta \mathbf{c}$ for modifications from female to male. The results for the remaining occupations are provided in the supplementary materials.

\subsubsection{Transition Score}
\label{ts}
Similar to Recall from FACET~\cite{gustafson2023facet},  we define the proportion of these altered outputs as the \textit{Transition Score}, which quantifies the extent of adjustment during debiasing. It is defined as:
\begin{equation}
\text{TS} = 
\begin{cases}
\frac{\#(\text{male} \rightarrow \text{female})}{\#(\text{original male})}, & \text{if reducing male bias}, \\
\frac{\#(\text{female} \rightarrow \text{male})}{\#(\text{original female})}, & \text{if reducing female bias}.
\end{cases}
\notag
\end{equation}

Unlike previous studies that apply debiasing uniformly across all generated images, our approach enables pre-specifying the proportion of samples to be debiased. As shown in Table~\ref{tab:occupation_all}, the \textsc{Transition Score (TS)} for most occupations closely align with the target ratio of 0.5. To explain, the original bias ratio $\Delta_{\text{org}}$ for the occupation \textit{doctor} is 0.617, corresponding to approximately 80.85\% male representations. When applying \textbf{DebFilter} with a debiasing ratio of 50\%, the expected \textsc{TS} is computed as $(80.85/100 \times 50)/80.85 = 0.5$, which closely approximates the observed value of 0.53. This demonstrates that \textbf{DebFilter} not only achieves approximate gender parity but also allows for flexible and controlled adjustment of the debiasing ratio. 

\begin{table}[tbp]
\centering
\footnotesize
\vspace{4pt}
\begin{tabular}{lrrrrrr}
\toprule
\textbf{Method} & Base & DeAR & Cl-clip & P-Deb & SFID & DebFilter \\
\midrule
\textsc{Skew} & 83.25 & 99.88 & 82.05 & 82.77 & 81.57 & \textbf{62.10} \\
\bottomrule
\end{tabular}
\caption{Neutral prompt score (↓ lower is better). The methods include: Baseline, Debiasing via Regularization (DeAR), 
CLIP-clip (Cl-clip), Prompt-Debias (P-Deb), Selective Feature Injection with Disentanglement using a hard constraint (SFID), and DebFilter.}
\label{tab:skew}
\end{table}

\subsubsection{Skew} SFID~\cite{jung2024unified} measured a fairness of generative models to produce an equal number of images for each gender across all occupations. Accordingly, the bias metric quantifies the degree of \textsc{Skew} in the gender distribution of the generated outputs:
{\small
\begin{equation}
\text{\textsc{Skew}} = \frac{1}{|P|} \sum_{p \in P} \frac{\max(N_{p,m}, N_{p,f})}{C}
\end{equation}
}
where \( N_{p,m} \) and \( N_{p,f} \) denote the number of detected male and female images, respectively, for occupation \( p \), and \(C\) is the number of image generations per prompt. In an ideal case with no bias, the \textsc{Skew} value would be 50.

Table~\ref{tab:skew} reports the \textsc{Skew} values across various debiasing methods, measuring the degree of gender imbalance in the generated images for a set of occupation prompts. Lower values indicate fairer gender distributions. While most methods show moderate improvements or even regressions compared to the baseline, \textbf{DebFilter} achieves the lowest \textsc{Skew} score of 62.1, demonstrating its effectiveness in mitigating gender bias across prompts. The results of our study were compared with the baseline results reported in SFID. For a fair comparison with recent studies, we use SDXL~\cite{podell2023sdxl} as the baseline diffusion model when computing $\Delta_p$.
\subsubsection{CLIPScore}
\begin{figure}[tbp]
    \centering
    \includegraphics[width=0.8\linewidth]{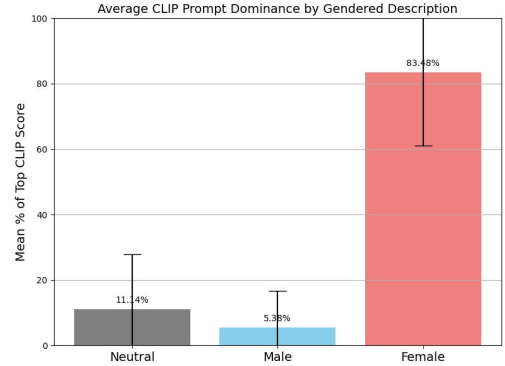}   \caption{Comparison of CLIPScores of female images evaluated with three types of prompts: a neutral prompt, a male-explicit prompt, and a female-explicit prompt.}
    \label{fig:clip-statistics}
\end{figure}
To further analyze the quality of debiased images, we calculate the \textsc{CLIPScore} for each image under different prompting conditions. Specifically, we use female-explicit prompts (e.g., ``A female sheriff'') when assessing \textsc{CLIPScore} for filtered images that have been transformed from male to female or vice versa.

We conduct an experiment using three types of prompts: a neutral prompt without any gender terms, a male-explicit prompt, and a female-explicit prompt. By comparing \textsc{CLIPScore}, we assess which type of prompt yields the highest alignment between the generated images and their corresponding text descriptions.

As can be seen in Figure~\ref{fig:clip-statistics}, in cases where the image has been debiased toward a female presentation, we observe that the \textsc{CLIPScore} is higher when using a female-explicit prompt. Conversely, images that retain a male presentation tend to achieve higher \textsc{CLIPScore} under the neutral prompt. This suggests that debiased images align more closely with gender-explicit language, while original male-presenting images remain better aligned with gender-neutral descriptions.

\textbf{DebFilter} is capable of effectively addressing the target bias without compromising the original semantic content or meaning of the generated outputs. As can be seen from Table~\ref{tab:occupation_all}(c), \textbf{DebFilter} achieves comparable or even higher scores compared to the \textsc{CLIPScore} obtained using the vanilla model, demonstrating that it produces results that more faithfully align with the given prompt.

\subsection{Age Bias Mitigation}
\label{sec:exp-age}
Broadly, age in bias mitigation area is classified into three categories: young (0–30), middle-aged (31–60), and elderly (60+). Certain words representing professions or individuals tend to be associated with either young or elderly age groups. We choose 14 words that refer to a person performing a specific action for evaluation(e.g., clock-maker). 
\begin{figure}[tbp]
    \centering
    \includegraphics[width=0.88\linewidth]{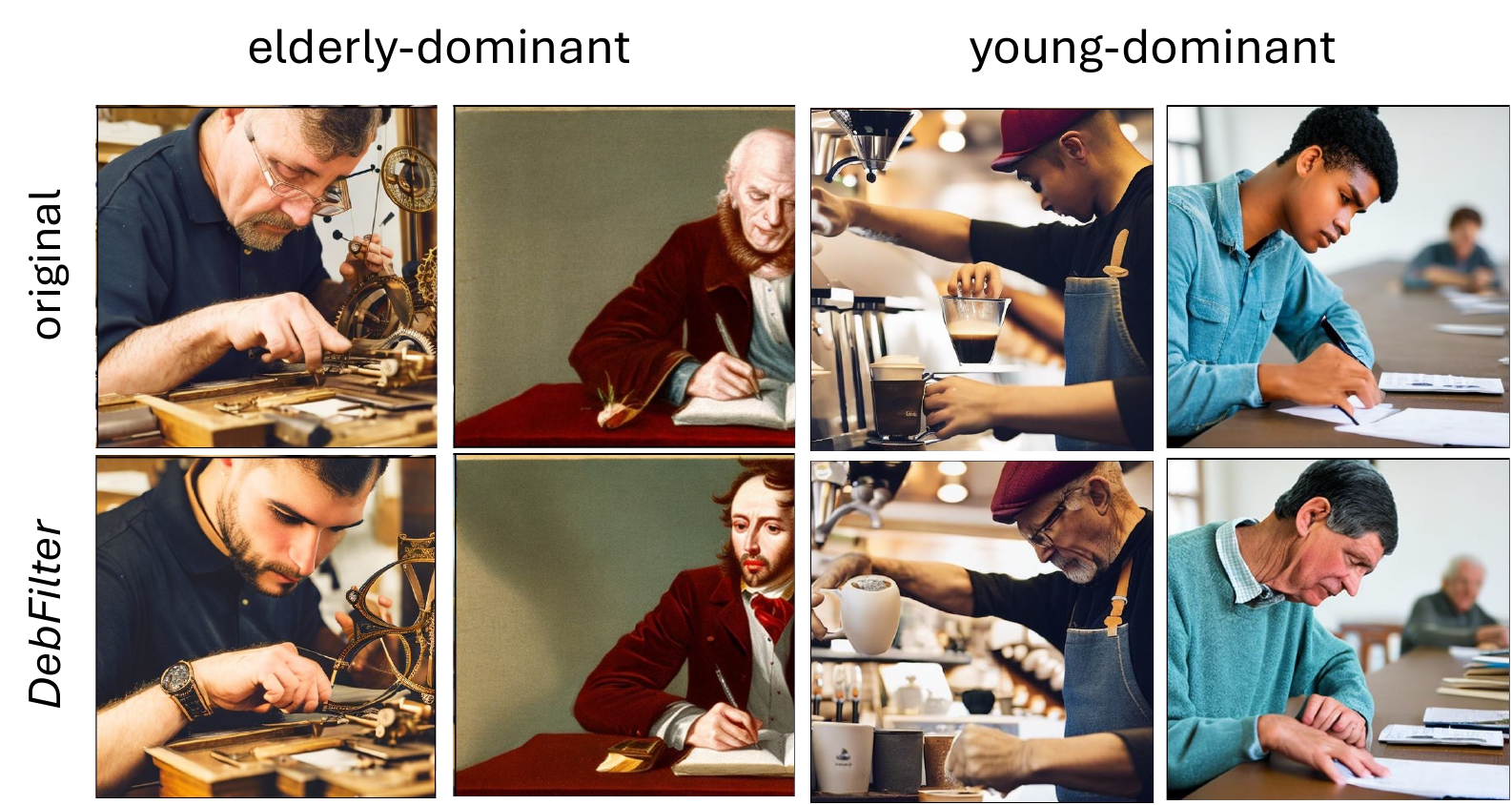}
    \caption{Examples of \textit{clock maker}, \textit{philosopher}, \textit{barista} and \textit{college student} in age bias mitigation using \textit{DebFilter}.}
    \label{fig:e2y}
\end{figure}
The samples in the top row of Figure \ref{fig:e2y} demonstrate that prompts containing certain professions such as ``clockmaker" or ``philosopher" exhibit a bias toward generating only older individuals while ``barista'' or ``student'' shows the opposite.
To make \textbf{DebFilter} for age, \( \Delta{\textbf{c}} \) is computed with target prompt that `\textit{old}' is inserted in front of the subject of source prompt(e.g., the target prompt of ``a barista making latte in a café'' becomes ``an old barista making latte in a café'').

In the same manner as described in debiasing female-dominant cases, the \textbf{DebFilter} for young-to-elderly transformation was stored and applied in the reverse direction. Figure~\ref{fig:e2y} shows that not only the appearance of a person such as hair style or wrinkles in skin is changed, style of the outfit also naturally changed compared to original samples. Though the action or social role is awkward with the age, \textbf{DebFilter} translates the semantic information naturally within the unchanged backgrounds.

\subsection{Multi-Object}
\label{sec:exp-multi}
\begin{figure}[tp]
    \centering
    \includegraphics[width=0.8\linewidth]{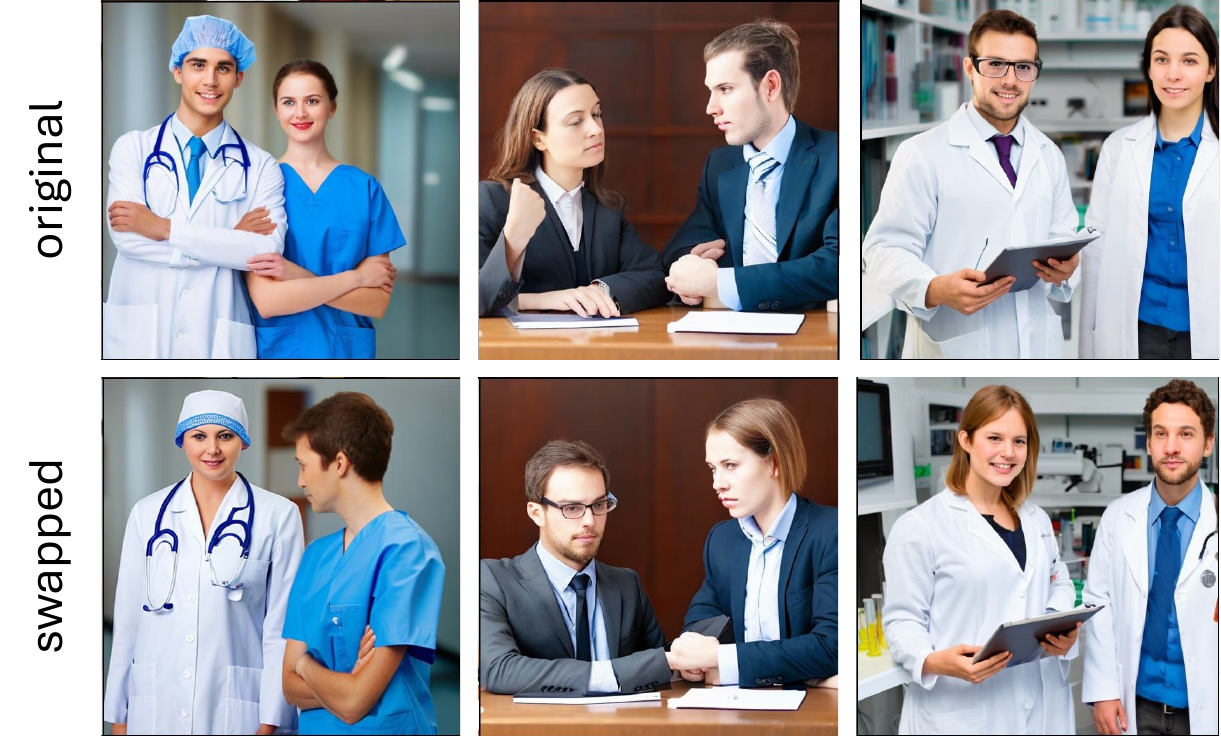}
    \caption{Gender-reversed examples of ``\textit{a doctor and a nurse}", ``\textit{a lawyer and a client}" and ``\textit{a scientist with an assistant}".}
    \label{fig:multi-object}
\end{figure}

Previous studies~\cite{chefer2023attend, li2023divide, hertz2022prompt} have focused on mitigating bias solely in the context of single objects. While TIME\cite{orgad2023editing} find it hard to control multiple professions at the same time, we further demonstrate that our proposed method can be applied to multi-object scenarios as well. 

Using the prompt ``a doctor and a nurse,’’ we applied the male-to-female \textbf{DebFilter} to the doctor index and its inverse to the nurse index. As a result, the vector $\mathbf{v}$ projected from the \textbf{DebFiltered} embedding for a male-dominant concept (e.g., ``doctor’’) is shifted toward the female direction, while the $\mathbf{v}$ associated with a female-dominant concept (e.g., ``nurse’’) is adjusted in the opposite direction.

The examples in the lower row of Figure~\ref{fig:multi-object} show that the original male depiction (upper row) is transformed to appear female, and conversely, the original female depiction is shifted toward male presentation. Importantly, these transformations preserve the overall layout and background structure of the images. Additional examples across diverse object pairs are provided in the supplementary materials.

%% file: sec/5_conclusion.tex
\section{Conclusion}
In this work, we explored how subtle perturbations to textual guidance can reshape the score-function landscape in diffusion-based generative models and leveraged this property to address social bias. We propose \textbf{DebFilter}, a lightweight and training-free framework that mitigates biases—particularly related to gender and age—by adjusting guidance embeddings. Operating at inference time, \textbf{DebFilter} offers fine-grained and interpretable control over debiasing strength while preserving semantic fidelity and image quality. Experimental results demonstrate that it effectively reduces bias across multiple metrics without retraining or modification. This shows that controlled guidance adjustment provides a powerful and scalable approach toward fairer and inclusive text-to-image generation. \textbf{DebFilter} can be extended to address other bias concepts such as ethnicity or cultural concepts.

\paragraph{Acknowledgment} This work was supported in part by Institute of Information \& communications Technology Planning \& Evaluation (IITP) grant funded by the Ministry of Science and ICT (MSIT), Korea Government (No. RS-2024-00428780). 

%% file: sec/X_suppl.tex
\clearpage
\setcounter{page}{1}

\maketitlesupplementary

\section{Configuration of Denoising U-Net}
\begin{table}[ht]
\centering
\small{
\begin{tabular}{c|cccccccc}
\toprule
\textbf{Index of CA} & 1 & 2 & 3 & 4 & 5 & 6 & 7 & 8 \\
\midrule
\textbf{Number of Heads} & 5 & 5 & 10 & 10 & 20 & 20 & 20 & 20 \\
\textbf{Dimension}       & 320 & 320 & 640 & 640 & 1280 & 1280 & 1280 & 1280 \\
\midrule
\textbf{Index of CA} & 9 & 10 & 11 & 12 & 13 & 14 & 15 & 16 \\
\midrule
\textbf{Number of Heads} & 20 & 20 & 10 & 10 & 10 & 5 & 5 & 5 \\
\textbf{Dimension}       & 1280 & 1280 & 640 & 640 & 640 & 320 & 320 & 320 \\
\bottomrule
\end{tabular}
}
\vspace{4pt}
\caption{Cross-attention (CA) configurations used in the denoising U-Net. The dimension of single head is set as 64.}
\label{tab:ca_config}
\end{table}

Table~\ref{tab:ca_config} summarizes the cross-attention (CA) configurations used across the denoising U-Net. The CA modules follow the standard U-shaped architecture, where attention capacity increases as depth increases in the encoder and decreases symmetrically during decoding. The guidance embedding of dimension 1024 is linearly projected to the layer-specific value dimensions listed in the table. Encoder depths 1--3 (CA~1--6) progressively expand the number of attention heads and value dimensions, the bottleneck (CA~7) serves as the highest-capacity point, and decoder depths 3--1 (CA~8--16) mirror this structure in reverse.

This structural pattern is directly connected to our unified guidance estimation method described in Sec. 3.2. Because each layer has its own projection matrix $W_v^{l,h}$ and operates in a different value-space determined solely by the architecture, the per-layer target vectors $\hat{\mathbf{v}}_{m}^{\,l,h}$ naturally lie in heterogeneous feature spaces. Our method explicitly leverages this architectural property: we estimate the targets at every layer and head independently, and then unify them via a weighted least-squares regression over all $W_v^{l,h}$. Thus, the hierarchical CA configuration provides the structural background that motivates aggregating multi-depth cross-attention outputs into a single coherent embedding~$\mathbf{c}^{\text{new}}_m$.

\section{Generalizability of DebFilter}
\begin{figure}[htb]
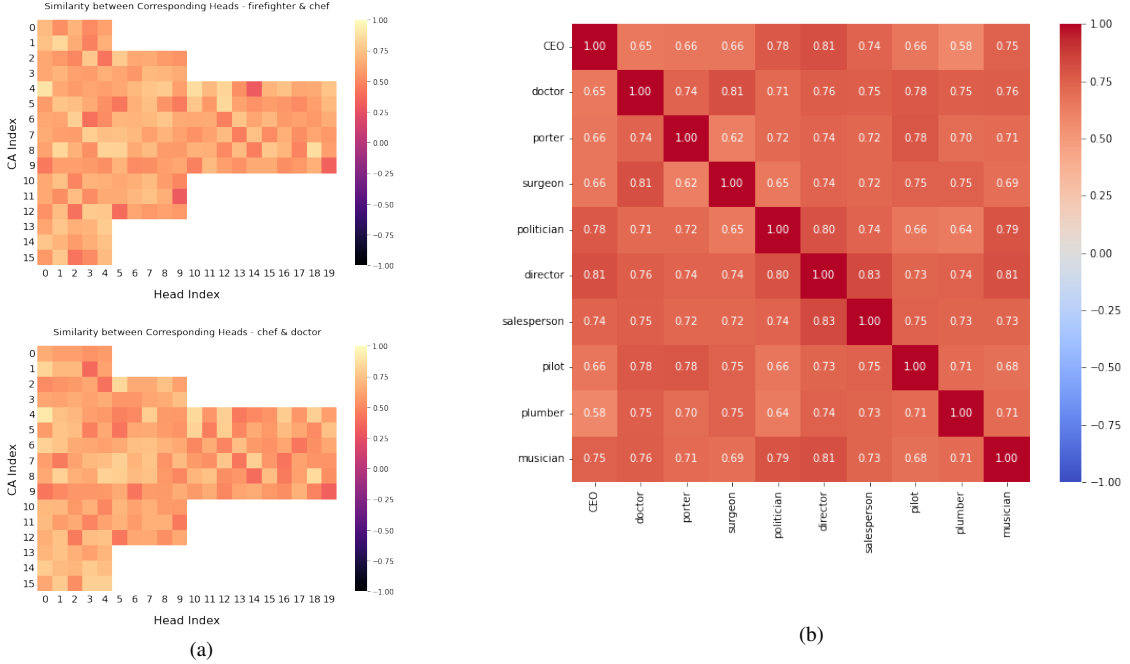

    \centering
    \begin{minipage}[c]{0.4\textwidth}
        \centering
        \includegraphics[width=0.75\linewidth]{fig/firefighter-chef.pdf}\\[6pt]
        \includegraphics[width=0.75\linewidth]{fig/chef-doctor.pdf}
        \subcaption{}
        \label{fig:head_comparison}
    \end{minipage}
    \hspace{6pt}
    \begin{minipage}[c]{0.48\textwidth}
        \centering
        \includegraphics[width=\linewidth]{fig/cosine_similarity.pdf}
        \vspace{9pt}
        \subcaption{}
        \label{fig:cosine_similarity}
    \end{minipage}
    \caption{
        Generalizability of representation shifts across occupations.
        (a) Cosine similarity between corresponding attention heads across layers for occupation pairs,
        comparing vectors $\Delta v_m^h$ for each head. Both analyses indicate that the model's occupation-specific changes are highly consistent across occupations.
        (b) Cosine similarity between normalized 1024-dimensional $\Delta c$ vectors across occupations, showing that many occupations share closely aligned shift directions.
    }
    \label{fig:generalizability}
\end{figure}
To evaluate whether a \textbf{DebFilter} extracted from one occupation can generalize to others, we investigate two complementary levels of directional consistency: (1) vector-level alignment in the value space, and (2) consistency within the guidance embeddings themselves.

\subsection{Vector-Level Comparison}

We first analyze the offsets arising in the value vectors. For a neutral prompt of the form \textit{``a person who works as \{occupation\}''}, we compute the modified value vector $\hat{v}_m$ that matches the cross-attention output obtained under an explicitly gender-specified prompt. The resulting offset
\[
\Delta v_m = \hat{v}_m - v_m
\]
captures the gender-associated adjustment at token index $m$. Since each transformer layer contains multiple attention heads, we denote by $\Delta v_m^h$ the offset corresponding to head $h$ within a given layer.

Following the configuration in Table~\ref{tab:ca_config}, we collect all $\Delta v_m^h$ across layers and heads for each occupation and examine their cross-occupation similarity. We find that offsets originating from the same layer and head are highly aligned across occupations, typically exhibiting cosine similarity values above $0.7$. This strong alignment suggests that the extracted direction is not unique to a specific occupation; rather, it reflects a shared latent gender-related semantic direction encoded consistently within the value space.

As illustrated in Figure~\ref{fig:head_comparison}, the projected offsets obtained from one occupation (e.g., \textit{chef}) closely match those derived from others such as \textit{doctor} and \textit{firefighter}. This cross-occupation coherence provides empirical evidence that the value-space debiasing direction expressed by \textbf{DebFilter} is broadly transferable, supporting both the robustness and generality of our method.

\subsection{Guidance-Level Comparison}

We further assess generalizability by examining consistency in the guidance embeddings. The $m$-th slice of the computed guidance embedding $\hat{\mathbf{c}}_m$ is a 1024-dimensional vector. When comparing $\hat{\mathbf{c}}_m$ across occupations, we observe consistently high cosine similarity scores, indicating strong directional alignment among the induced representation shifts.

To quantify this effect, each $\Delta c_m$ vector is first normalized, and pairwise cosine similarities are computed across occupations. As shown in Figure~\ref{fig:cosine_similarity}, the resulting heatmap demonstrates uniformly elevated similarity values (typically in the 0.7–0.85 range), revealing that the shift directions in guidance space cluster tightly around a shared axis. This pattern indicates that the model applies nearly parallel adjustment directions across a wide variety of occupations. Such strong cross-occupation alignment highlights that \textbf{DebFilter} relies on a stable, occupation-invariant mechanism when encoding gender-related corrections, further reinforcing its generalizability beyond the specific occupation from which it was derived.

\section{Sensitivity in Constructing DebFilter}

\begin{figure*}[t]
    \centering

    \begin{subfigure}{0.45\linewidth}
        \centering
        \includegraphics[width=\linewidth]{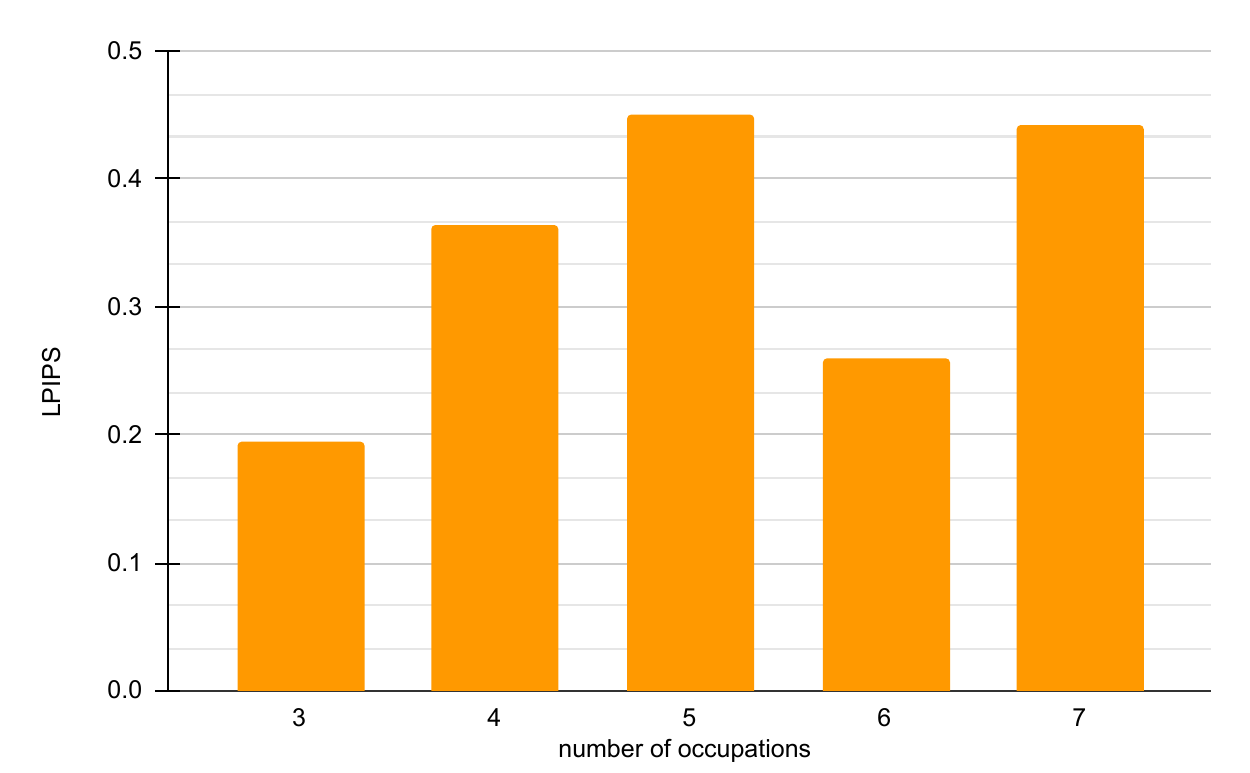}
        \caption{LPIPS vs. number of occupations used for averaging.}
        \label{fig:lpips_a}
    \end{subfigure}
    \hspace{10pt}
    \begin{subfigure}{0.45\linewidth}
        \centering
        \includegraphics[width=\linewidth]{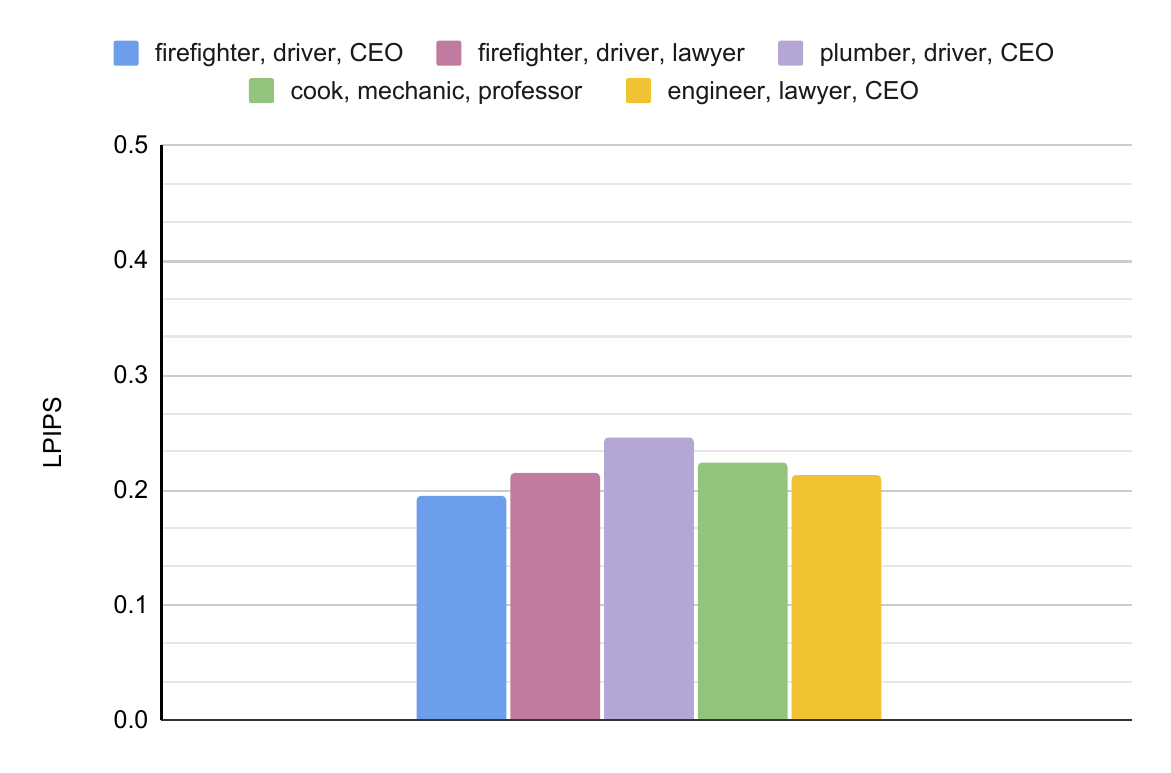}
        \caption{LPIPS across different occupation subsets.}
        \label{fig:lpips_b}
    \end{subfigure}

    \vspace{10pt}
    \begin{subfigure}{0.8\linewidth}
        \centering
        \includegraphics[width=0.75\linewidth]{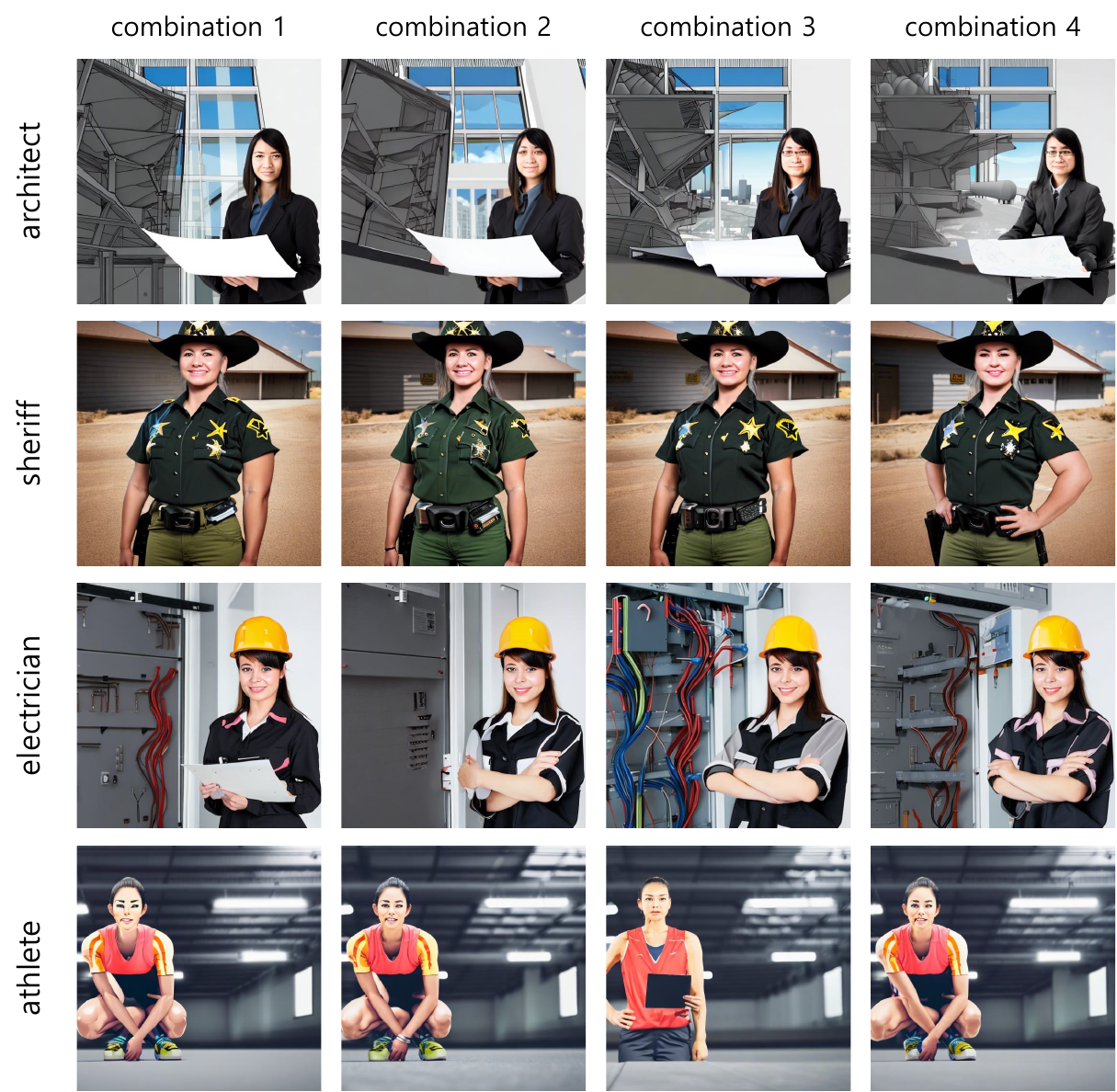}
        \caption{Sensitivity of \textbf{DebFilter} to the choice of occupations. Different occupation subsets yield visually similar outputs, indicating stable debiasing behavior.}
        \label{fig:sense_ex}
    \end{subfigure}

    \caption{
        Summary of \textbf{DebFilter} sensitivity analysis. 
        (a,b) LPIPS evaluation showing that using three occupations produces the lowest perceptual distortion, and the specific choice of occupations has only minor impact. 
        (c) Debiased image examples across multiple target occupations. 
        (d) Sensitivity visualization demonstrating minimal visual variation when \textbf{DebFilter} is constructed from different occupation subsets.
    }
    \label{fig:sensitivity_summary}
\end{figure*}

    

To construct the \textbf{DebFilter}, we first evaluated how each occupation-specific $\Delta c_m$ behaves when applied to its corresponding occupation. This analysis helped identify which occupations provide stable and representative debiasing signals. While several occupations produced reasonable results, the combination of \textit{firefighter}, \textit{driver}, and \textit{CEO} showed the most consistent behavior and was therefore used to compute the averaged transformation.

As shown in Figure~\ref{fig:lpips_a}, using three occupations yields the lowest LPIPS score, indicating that averaging over three well-chosen occupations is sufficient---and even optimal---for minimizing perceptual distortion introduced by the debiasing direction. Increasing the number of occupations beyond three does not improve generalization; rather, it tends to introduce additional noise, resulting in higher LPIPS values.

Crucially, the overall debiasing performance is not sensitive to the specific occupations chosen for constructing the \textbf{DebFilter}. As shown in Figure~\ref{fig:lpips_b}, substituting the original three occupations with other plausible subsets of comparable size leads to only minor fluctuations in LPIPS. This invariance is further illustrated in Figure~\ref{fig:sense_ex}, where the resulting generated images remain visually consistent across different occupation subsets. These results indicate that \textbf{DebFilter} is effectively agnostic to choice of the occupations used in construction. Thus, a moderately sized subset—typically three occupations—is both sufficient and robust, eliminating the need for precise or carefully curated selection.

\section{Isolating Central Traits}
\begin{figure}[!htbp]
    \centering
    \includegraphics[width=0.5\linewidth]{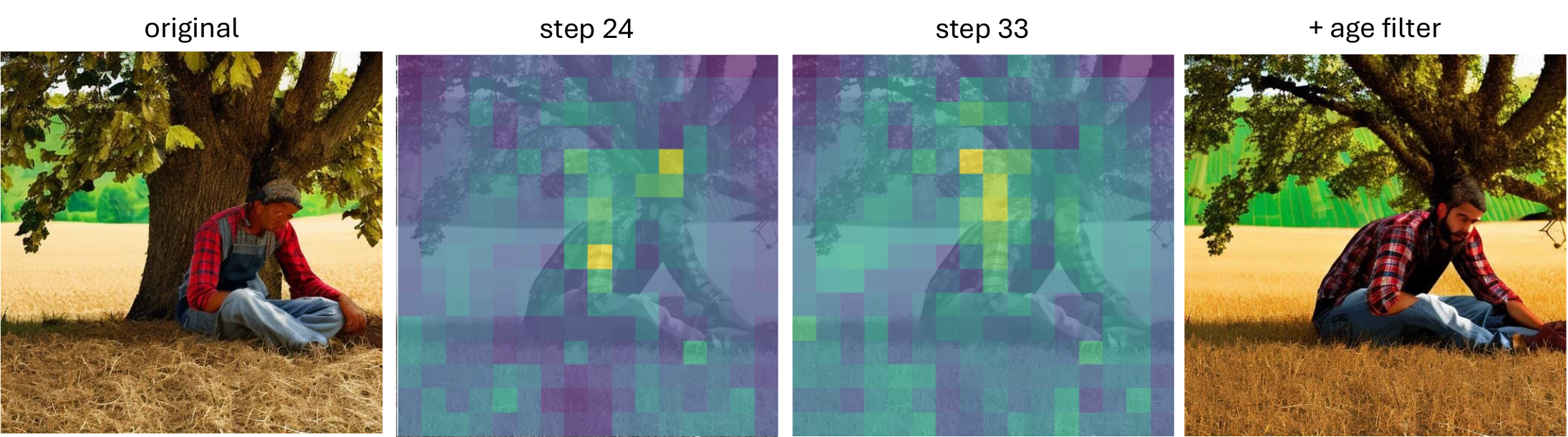} \\
    \includegraphics[width=0.5\linewidth]{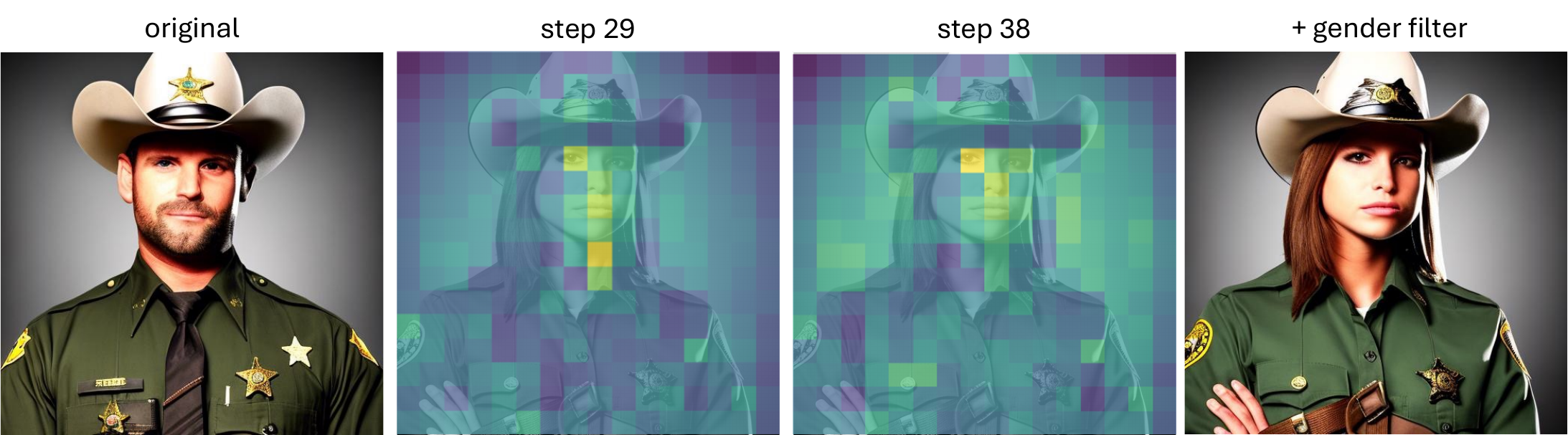}
    \caption{Examples of attention maps during denoising for ``farmer'' and ``sheriff.'' From left to right: the original image, attention-difference maps at intermediate steps, and the final output after applying the \textbf{DebFilter}. Brighter regions show where $\Delta v_m$ induces the largest changes, primarily on person-related areas, indicating targeted concept editing.}
    \label{fig:attnmap}
\end{figure}
The images on the right illustrate the outcomes obtained by applying \textbf{DebFilter} with different bias-related concepts. To analyze how \textbf{DebFilter} modifies internal representations, we visualize the corresponding attention score differences by multiplying the cross-attention maps with the \textbf{DebFilter}-derived vector shift, $\Delta v_m$. Brighter regions indicate locations where the modification induced by $\Delta v_m$ is most pronounced, and these regions spatially overlap with areas of the original image that are semantically linked to the targeted bias concept. This confirms that simply modifying the $m$-th token by adding $\Delta v_m$ primarily alters the intended concept while leaving unrelated regions largely unaffected.

Since semantic structure in diffusion models becomes increasingly stable after approximately half of the denoising trajectory, we visualize representative steps after step~25, where conceptual information is well formed. As shown in Figure~\ref{fig:attnmap}, \textbf{DebFilter} produces debiasing effects while preserving global scene layout and background consistency. In both examples, the model modifies only the targeted attributes (e.g., age or gender) while maintaining all unrelated visual features.

\textbf{DebFilter} was derived using a range of occupations and even when unrelated professions are incorporated during the averaging process, no occupational artifacts appear in the results. This demonstrates that \textbf{DebFilter} operates directly within the attention mechanism, enabling precise concept-level editing without leaking unintended semantics.
\newpage
\section{Qualitative Results of Gender Debiasing}
\begin{figure*}[htbp]
    \centering
    \begin{subfigure}{0.48\textwidth}
        \includegraphics[width=\linewidth]{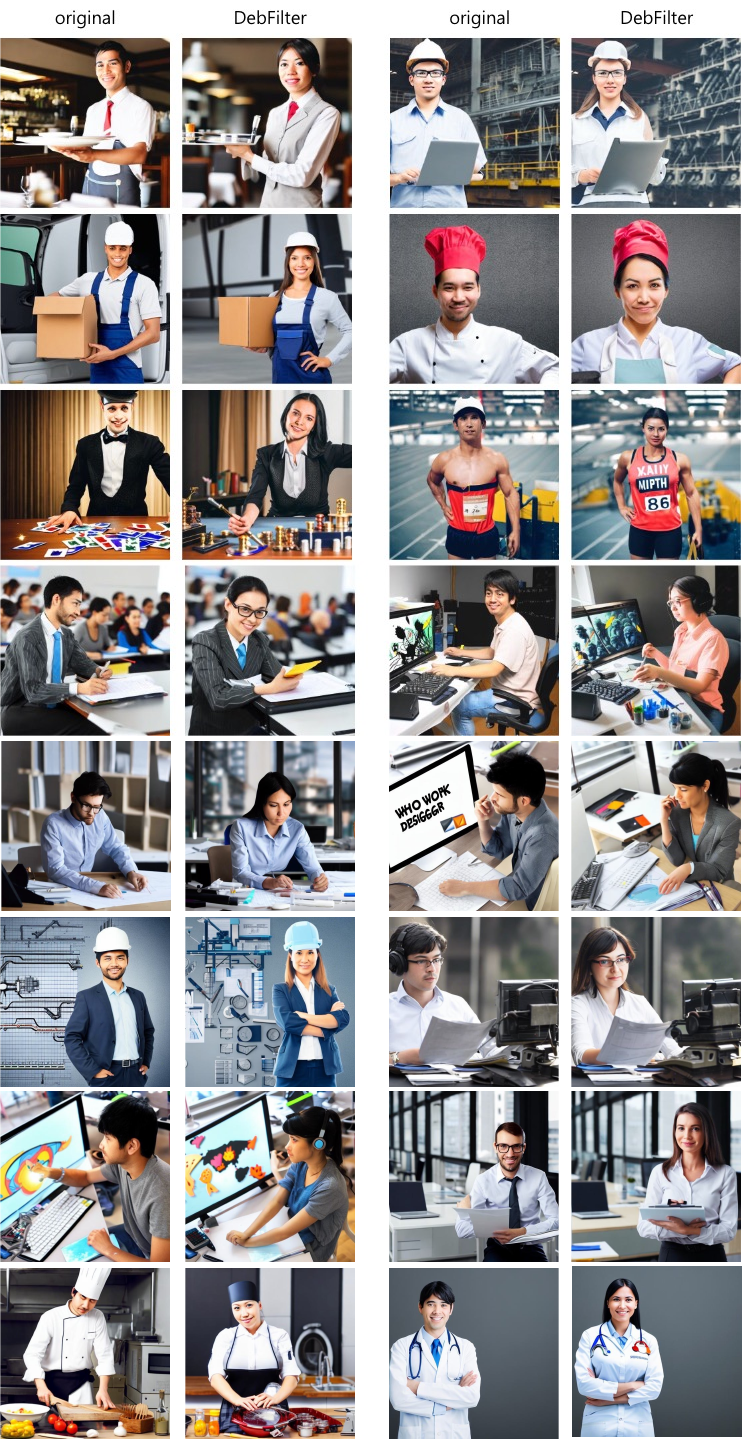}
        \caption{Male-to-Female}
        \label{fig:male-to-female}
    \end{subfigure}
    \hfill
    \begin{subfigure}{0.48\textwidth}
        \includegraphics[width=\linewidth]{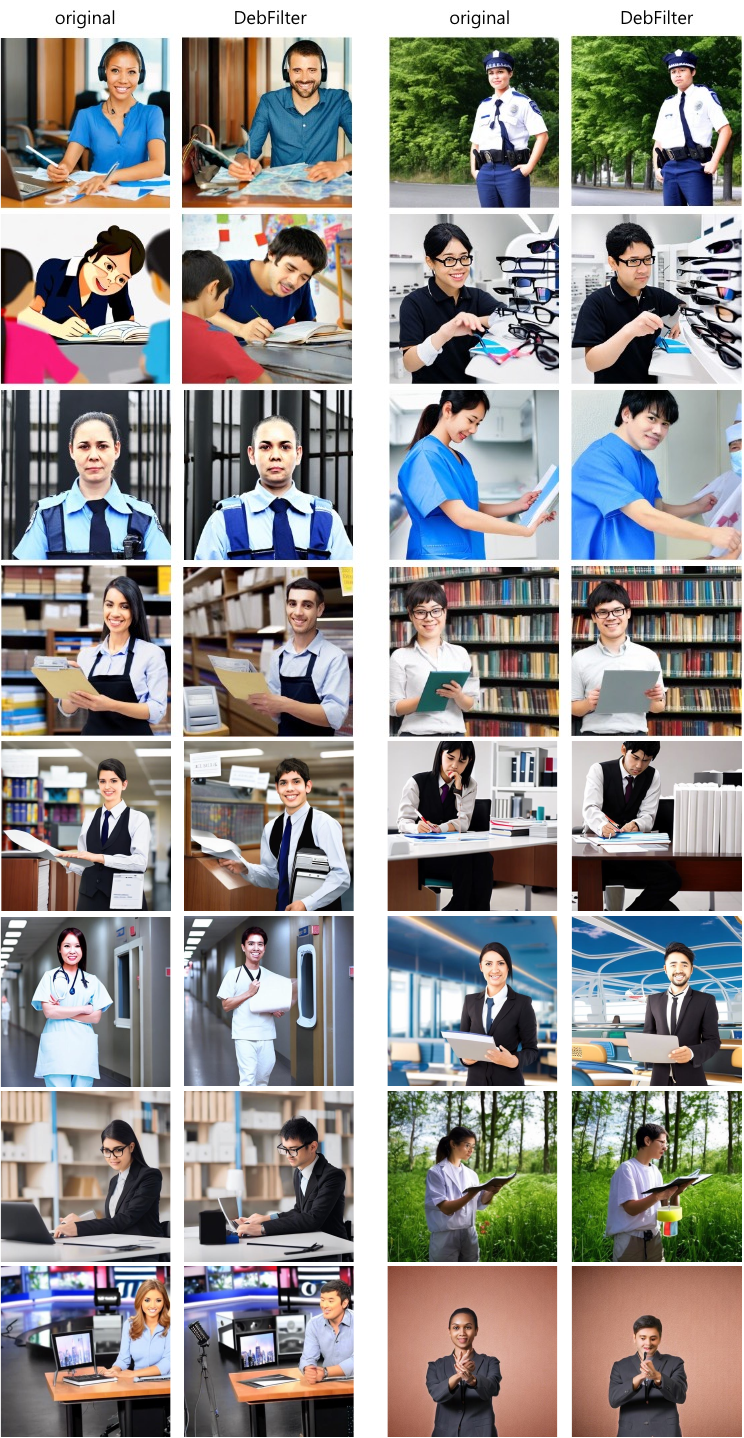}
        \caption{Female-to-Male}
        \label{fig:female-to-male}
    \end{subfigure}
    \caption{Illustration of gender transformation tasks.}
    \label{fig:gender-transformation}
\end{figure*}
\clearpage

\clearpage
\section{Quantitative Results in Bias Ratio male-to-female and female-to-male debiasing}
\begin{table*}[htbp]
\centering
\caption{$\Delta_{org}$ and $\Delta_{deb}$ values across occupations (male-to-female and female-to-male).}
\footnotesize
\setlength{\tabcolsep}{4pt}
\begin{tabular}{lcc lcc lcc}
\toprule
\multicolumn{3}{c}{\textbf{Male-to-Female}} & \multicolumn{3}{c}{\textbf{Male-to-Female}} & \multicolumn{3}{c}{\textbf{Female-to-Male}} \\
\cmidrule(r){1-3} \cmidrule(lr){4-6} \cmidrule(l){7-9}
Occupation & $\Delta_{org}\downarrow$ & $\Delta_{deb}\downarrow$ & Occupation & $\Delta_{org}\downarrow$ & $\Delta_{deb}\downarrow$ & Occupation & $\Delta_{org}\downarrow$ & $\Delta_{deb}\downarrow$ \\
\midrule
magician          & 0.983 & 0.000 & electrician      & 0.983 & 0.033 & accountant           & 0.100 & 0.567 \\
pilot             & 0.733 & 0.000 & plumber          & 0.983 & 0.033 & assistant            & 0.350 & 0.300 \\
architect         & 0.900 & 0.017 & trucker          & 0.983 & 0.033 & auditor              & 0.033 & 0.500 \\
sheriff           & 0.900 & 0.017 & film director    & 0.933 & 0.033 & baker                & 0.200 & 0.550 \\
executive         & 0.883 & 0.017 & butcher          & 0.900 & 0.033 & biologist            & 0.350 & 0.283 \\
professor         & 0.867 & 0.017 & mechanic         & 1.000 & 0.050 & career counselor     & 0.083 & 0.400 \\
director          & 0.800 & 0.017 & economist        & 0.933 & 0.050 & caretaker            & 0.467 & 0.200 \\
journalist        & 0.783 & 0.017 & editor           & 0.833 & 0.067 & civil servant        & 0.183 & 0.300 \\
manager           & 0.683 & 0.017 & lawyer           & 0.833 & 0.067 & cleaner              & 0.650 & 0.767 \\
writer            & 0.633 & 0.017 & scientist        & 0.783 & 0.067 & clerk                & 0.600 & 0.250 \\
veterinarian      & 0.517 & 0.017 & musician         & 0.867 & 0.100 & dentist              & 0.317 & 0.367 \\
electrician       & 0.983 & 0.033 & chef             & 0.833 & 0.100 & hairdresser          & 0.317 & 0.417 \\
plumber           & 0.983 & 0.033 & garbage collector& 0.983 & 0.133 & housekeeper          & 0.950 & 0.950 \\
trucker           & 0.983 & 0.033 & cook             & 0.817 & 0.117 & jeweler              & 0.883 & 0.950 \\
film director     & 0.933 & 0.033 & photographer     & 0.650 & 0.100 & library assistant    & 0.883 & 0.067 \\
butcher           & 0.900 & 0.033 & decorator        & 0.900 & 0.183 & makeup artist        & 1.000 & 0.383 \\
mechanic          & 1.000 & 0.050 & designer         & 0.733 & 0.150 & nurse                & 0.950 & 0.050 \\
economist         & 0.933 & 0.050 & company director & 0.800 & 0.167 & optician             & 0.467 & 0.017 \\
editor            & 0.833 & 0.067 & miner            & 1.000 & 0.233 & personal assistant   & 0.817 & 0.017 \\
lawyer            & 0.833 & 0.067 & salesperson      & 0.467 & 0.117 & police officer       & 0.417 & 0.133 \\
scientist         & 0.783 & 0.067 & animator         & 0.783 & 0.200 & prison officer       & 0.200 & 0.367 \\
musician          & 0.867 & 0.100 & guard            & 0.833 & 0.217 & receptionist         & 0.950 & 0.867 \\
chef              & 0.833 & 0.100 & comicbook writer & 0.383 & 0.100 & sailor               & 0.717 & 0.067 \\
garbage collector & 0.983 & 0.133 & mover            & 1.000 & 0.267 & secretary            & 1.000 & 0.867 \\
cook              & 0.817 & 0.117 & athlete          & 0.933 & 0.283 & shop assistant       & 0.083 & 0.383 \\
photographer      & 0.650 & 0.100 & builder          & 1.000 & 0.367 & signlanguage interpreter & 0.533 & 0.250 \\
decorator         & 0.900 & 0.183 & politician       & 0.950 & 0.367 & teacher              & 0.383 & 0.233 \\
designer          & 0.733 & 0.150 & doctor           & 0.617 & 0.250 & translator           & 0.117 & 0.433 \\
company director  & 0.800 & 0.167 & soldier          & 0.517 & 0.217 & travel agent         & 0.567 & 0.183 \\
miner             & 1.000 & 0.233 & tailor           & 0.550 & 0.250 & tv presenter         & 0.633 & 0.183 \\
salesperson       & 0.467 & 0.117 & farmer           & 0.950 & 0.450 &                      &       &       \\
animator          & 0.783 & 0.200 & lecturer         & 0.500 & 0.283 &                      &       &       \\
guard             & 0.833 & 0.217 & surgeon          & 0.433 & 0.250 &                      &       &       \\
comicbook writer  & 0.383 & 0.100 & janitor          & 0.967 & 0.867 &                      &       &       \\
mover             & 1.000 & 0.267 & laborer          & 0.967 & 0.883 &                      &       &       \\
athlete           & 0.933 & 0.283 & geologist        & 0.950 & 0.883 &                      &       &       \\
builder           & 1.000 & 0.367 & waiter           & 0.250 & 0.433 &                      &       &       \\
politician        & 0.950 & 0.367 & singer           & 0.067 & 0.433 &                      &       &       \\
doctor            & 0.617 & 0.250 & solicitor        & 0.017 & 0.433 &                      &       &       \\
soldier           & 0.517 & 0.217 & porter           & 0.983 & 0.317 &                      &       &       \\
tailor            & 0.550 & 0.250 & painter          & 0.950 & 0.317 &                      &       &       \\
farmer            & 0.950 & 0.450 & judge            & 0.600 & 0.033 &                      &       &       \\
lecturer          & 0.500 & 0.283 & engineer         & 0.883 & 0.050 &                      &       &       \\
surgeon           & 0.433 & 0.250 &                  &       &       &                      &       &       \\
janitor           & 0.967 & 0.867 &                  &       &       &                      &       &       \\
laborer           & 0.967 & 0.883 &                  &       &       &                      &       &       \\
geologist         & 0.950 & 0.883 &                  &       &       &                      &       &       \\
waiter            & 0.250 & 0.433 &                  &       &       &                      &       &       \\
singer            & 0.067 & 0.433 &                  &       &       &                      &       &       \\
solicitor         & 0.017 & 0.433 &                  &       &       &                      &       &       \\
porter            & 0.983 & 0.317 &                  &       &       &                      &       &       \\
painter           & 0.950 & 0.317 &                  &       &       &                      &       &       \\
judge             & 0.600 & 0.033 &                  &       &       &                      &       &       \\
engineer          & 0.883 & 0.050 &                  &       &       &                      &       &       \\
\bottomrule
\end{tabular}
\label{tab:combined_gender_bias}
\end{table*}
Table~\ref{tab:combined_gender_bias} shows gender bias measurements for 60 occupations using two metrics: original bias (\(\Delta_{org}\) and bias after a debiasing intervention (\(\Delta_{debiased}\)). Values represent male-to-female associations, with higher numbers indicating stronger male stereotyping.

Many traditionally male-dominated occupations like magician, pilot, mechanic, and builder show very high original bias (0.983-1.000). After debiasing, most of these show dramatic reductions, often dropping to near zero. However, some occupations like janitor, laborer, and geologist retain substantial bias even after intervention (0.867-0.883).

Female-associated occupations like singer and solicitor start with low bias values (0.017-0.067) and show moderate increases after debiasing. The results suggest that while debiasing techniques can significantly reduce gender stereotypes in many professions, effectiveness varies considerably across different occupational categories.

In contrast to cases where male-associated terms dominate, we observed instances where, despite female-associated terms being closer in the embedding space, the generated outputs still tended to reflect male-related content more frequently or exhibited a more gender-balanced distribution. Consequently, by setting the ratio to 0.5, we examine the task of converting female to male representations and find that, as shown in Table~\ref{tab:combined_gender_bias}, the quality of the resulting \(\Delta\) improves in some cases, while in others, no such improvement is observed.\

Table~\ref{tab:combined_gender_bias} shows female-to-male gender bias scores \(\Delta_{org}\) and \(\Delta_{deb}\) across occupations before and after debiasing. Some roles, like ``makeup artist'' and ``secretary'', have high original bias, while others, like ``auditor," show low initial bias. Debiasing effectively reduces bias in many cases (e.g., ``library assistant''), but not all—``secretary'' and ``receptionist'' remain highly biased. In a few cases, bias increases after debiasing, highlighting that results vary by occupation and that debiasing methods may have inconsistent effects.
\section{Qualitative Results of Age Debiasing}
\begin{figure}[htbp]
    \centering
    \begin{subfigure}{0.45\textwidth}
        \includegraphics[width=\linewidth]{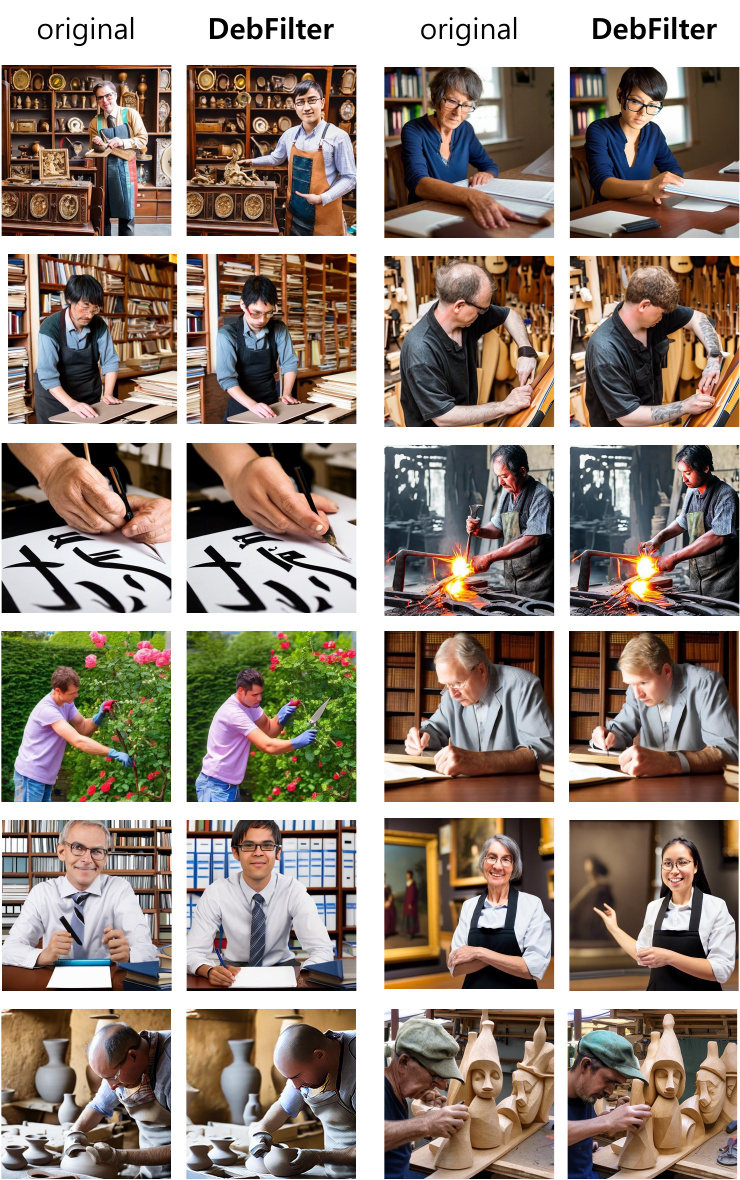}
        \caption{Elderly-to-Young}
        \label{fig:elderly-to-young}
    \end{subfigure}
    \hspace{2em} 
    \begin{subfigure}{0.45\textwidth}
        \includegraphics[width=\linewidth]{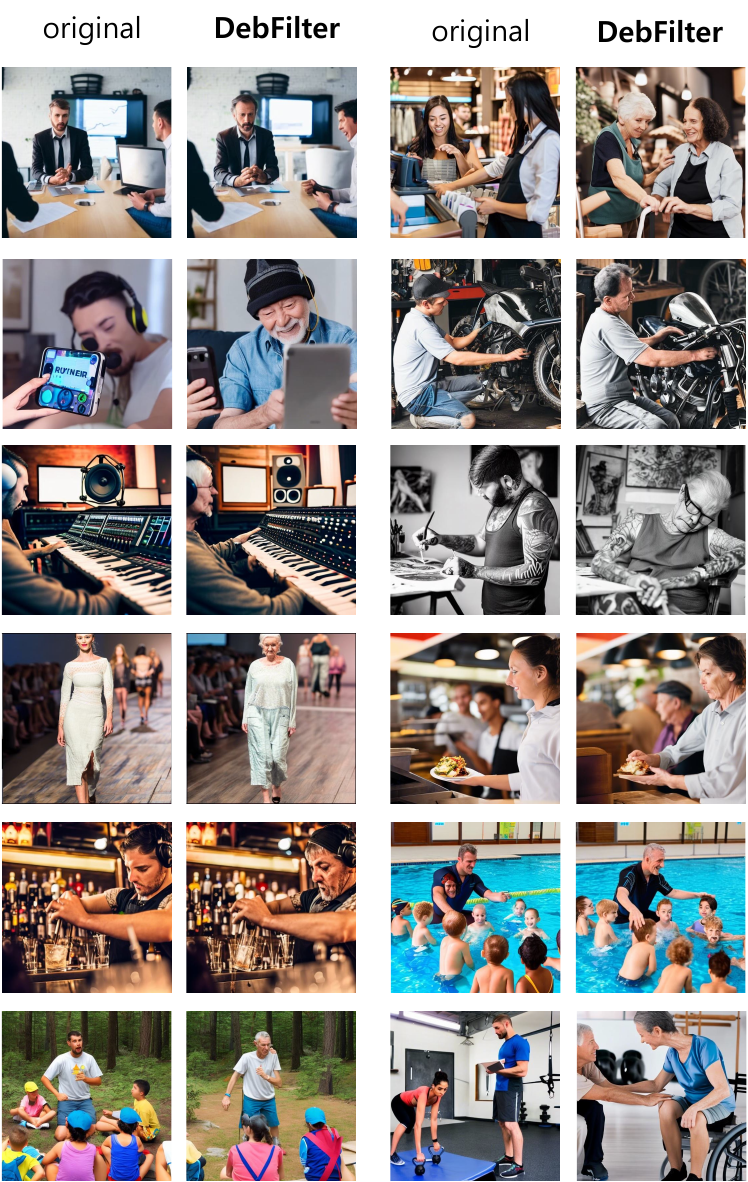}
        \caption{Young-to-Elderly}
        \label{fig:young-to-elderly}
    \end{subfigure}
    \caption{Illustration of age transformation tasks.}
    \label{fig:age-transformation}
\end{figure}
\clearpage

\section{Extension to Multi-Object Scenarios}
\begin{figure*}[htbp]
    \centering
    \includegraphics[width=0.85\textwidth]{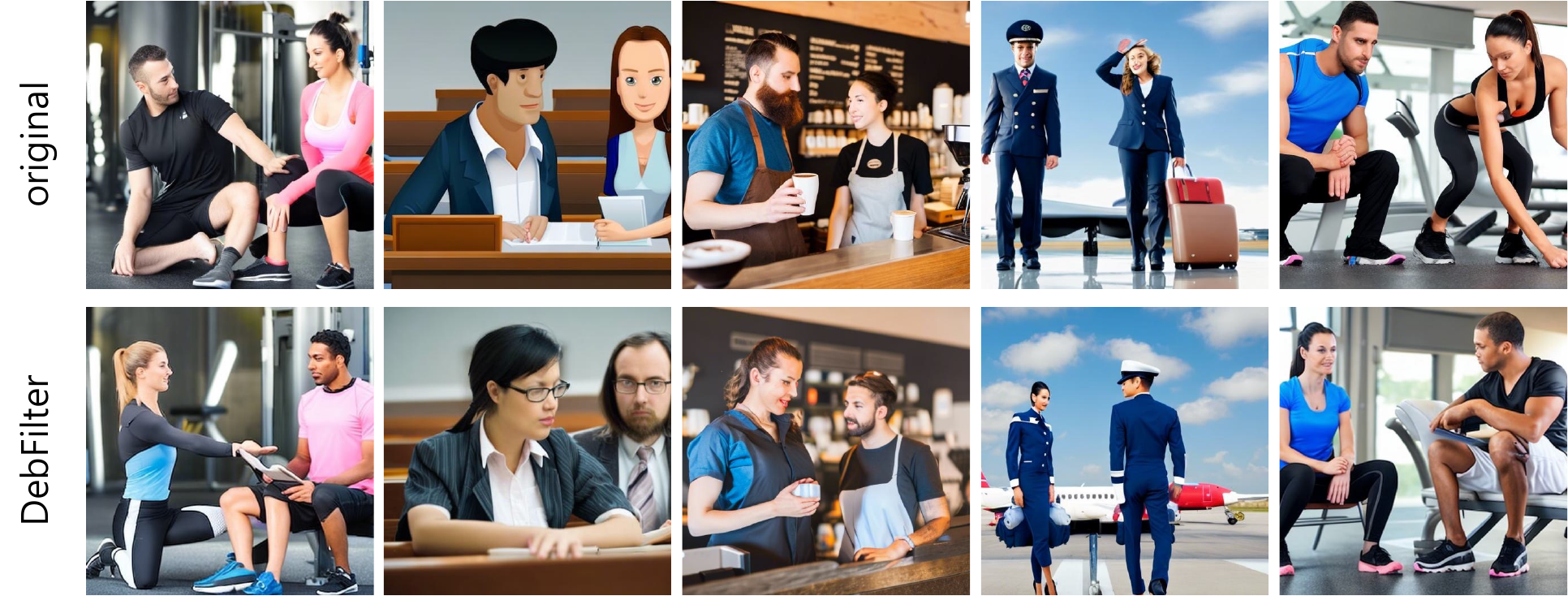}
    \caption{Localized gender transformation outcomes in multi-object image generation using \textbf{DebFilter}}
    \label{fig:more-multi}
\end{figure*}
A key limitation of vanilla Stable Diffusion is its tendency to misalign attributes with corresponding objects, often failing to represent semantic details such as gender or age correctly. Additionally, it struggles to generate coherent images from prompts involving multiple entities, limiting its effectiveness in complex, real-world scenarios. 

To address these limitations, recent studies have explored methods for improving object-attribute alignment and handling multi-entity prompts more effectively. These approaches aim to enhance controllability in text-to-image generation by incorporating structural guidance, grounding modules, or attention manipulation techniques.

Our proposed method builds on the observation that Stable Diffusion is capable of accurately rendering individual objects when their attributes are clearly described and disentangled. With Figure~\ref{fig:more-multi} we demonstrate that, under such conditions, our approach can effectively apply debiasing transformations to the intended target without affecting unrelated entities in multi-object scenarios.

\section{Drop-out Ratio}
\begin{table}[htbp]
    \centering
    \small
    \begin{tabular}{cc}
        \begin{subtable}{0.45\linewidth}
            \centering
            \begin{tabular}{cc|cc|c}
                & & \multicolumn{2}{c|}{debiased} & \\
                & & male & female & Total \\
                \hline
                \multirow{2}{*}{original} & male & 64 & 43 & 107 \\
                & female & 0 & 13 & 13 \\
                \hline
                & Total & 64 & 56 & 120\\
            \end{tabular}
            \caption{}
            \label{tab:gender-confusion}
        \end{subtable}
        &
        \begin{subtable}{0.45\linewidth}
            \centering
            \begin{tabular}{lc|lc}
                \toprule
                Occupation & \(\mathbf{F_{p}}\) & Occupation & \(\mathbf{F_{p}}\) \\
                \midrule
                farmer & 66.67 & manager & 66.67 \\
                mechanic & 66.67 & sheriff & 63.89 \\
                guard & 64.08 & cook & 66.67 \\
                lawyer & 63.89 & carpenter & 66.67 \\
                CEO & 61.11 & police officer & 65.42\\
                \bottomrule
            \end{tabular}
            \caption{}
            \label{tab:fp-values}
        \end{subtable}
    \end{tabular}
    \caption{(a) Comparison of original and debiased classification results by gender.  
    (b) Professions and their corresponding \(F_{p}\) values when the \textbf{DebFilter} ratio is 2/3.}
    \label{tab:combined-tables}
\end{table}

Table~\ref{tab:gender-confusion} illustrates the change in gender distribution for the prompt \textit{``A person who works as a surgeon''} after applying our proposed debiasing method. 
Using CLIP-based few-shot classification on images generated by the vanilla Stable Diffusion model, we initially observe 107 male-presenting and 13 female-presenting surgeons. 
After applying \textbf{DebFilter} with a target male-to-female ratio (\(p\)) of 0.5, 43 of the 107 originally male images are modified to appear female, resulting in a substantially more balanced distribution that aligns with the specified debiasing objective.

As shown in Table~\ref{tab:fp-values}, when the debiasing ratio is set to \(p = \frac{2}{3}\), the CLIP-based classification consistently yields a female proportion \(F_p\) of approximately 66\%. 
Similarly, a debiasing ratio of \(p = \frac{1}{3}\) produces an \(F_p\) near 33\%, while maintaining a stable overall gender bias score of \(\Delta \approx 0.36\). 
These results demonstrate that \textbf{DebFilter} enables precise and interpretable control over gender ratio adjustments without introducing instability or compromising consistency in the generated outputs.

It is important to note, however, that the resulting gender distribution does not always match the target ratio exactly. 
For instance, even when the debiasing ratio is set to \(p = \frac{2}{3}\) for occupations such as *CEO*, the generated female proportion may deviate slightly from the ideal value of $ 66.67\% $. 
This discrepancy arises because the baseline gender distribution produced by Stable Diffusion is not $100{:}0$; therefore, the model does not begin from a neutral or unbiased prior, and the applied adjustment interacts with this inherent distribution.

\section{Limitations}
\paragraph{Attribute Binding} However, one limitation arises from the inherent behavior of Stable Diffusion: when a prompt contains multiple parallel semantic components, the model often struggles to represent both attributes simultaneously with equal accuracy~\cite{chefer2023attend, li2023divide, hertz2022prompt}. 
If the base model fails to correctly associate an attribute with its intended subject, \textbf{DebFilter} may also inherit this limitation, causing the debiasing transformation to be applied disproportionately to a single object rather than being distributed across all relevant entities.

\paragraph{Discrepancy from the Real-World Ratio} According to the WinoBias taxonomy, certain occupations are socially male-dominant; however, in text-to-image diffusion models, the learned distribution sometimes exhibits a reversed tendency, generating more female-presenting images for these professions (e.g., \textit{singer}, \textit{waiter}, \textit{baker}, \textit{optician}). 
Such cases introduce misalignment between the expected debiasing direction and the model’s inherent bias pattern, which can complicate the application of a consistent gender-debiasing transformation.

\section{Compute Resources Used for Experiments}
\begin{table}[ht]
\centering
\caption{Compute Resources Used for Experiments}
\begin{tabular}{l|l}
\hline
\textbf{Component}           & \textbf{Specification}                     \\ \hline
\midrule
CPU                 & AMD Ryzen Threadripper PRO 3955WX \\ \hline
CPU Cores           & 16                                         \\ \hline
GPU Model           & NVIDIA RTX A5000                            \\ \hline
\end{tabular}
\end{table}